\documentclass[runningheads]{llncs}
\usepackage{graphicx}
\usepackage{amsmath,bm,amsfonts,amssymb,mathrsfs}
\usepackage{color}

\usepackage[utf8]{inputenc}
\usepackage[T1]{fontenc}
\usepackage[bookmarks=false]{hyperref}
\usepackage{url}
\usepackage{booktabs}
\usepackage{nicefrac}
\usepackage{microtype}
\usepackage{algorithmic}
\usepackage{bm}
\usepackage{grffile}
\usepackage{xspace}
\usepackage{etoolbox}

\hypersetup{colorlinks=true,hidelinks,bookmarks=false}

\newcommand{\eg}{\textit{e.g.}\xspace}
\newcommand{\ie}{\textit{i.e.}\xspace}

\newcommand{\etc}{\textit{etc.}\xspace}
\newcommand{\etal}{\textit{et~al.}\xspace}

\newcommand{\mathbold}[1]{\bm{#1}}
\newcommand{\mbf}[1]{\mathbf{#1}}
\newcommand{\vect}[1]{\mathbf{#1}}

\newcommand{\T}{^\mathsf{T}}

\newcommand{\vtheta}[0]{\mathbold{\theta}}

\newcommand{\vphi}[0]{\mathbold{\phi}}

\newcommand{\vx}{\mbf{x}}

\newcommand{\vz}{\mbf{z}}

\newcommand{\MI}{\mbf{I}}

\newcommand{\CelebA}{\textsc{CelebA}\xspace}
\newcommand{\CelebAHQ}{\textsc{CelebA-HQ}\xspace}
\newcommand{\Cifar}{\textsc{Cifar-10}\xspace}

\newcommand{\PIONEER}{\textsc{Pioneer}\xspace}

\usepackage{subcaption}

\usepackage{tikz,pgfplots,pgfmath}
\usetikzlibrary{plotmarks,arrows,calc,decorations.text}

\newlength{\figurewidth}
\newlength{\figureheight}

\newcommand{\toptitlebar}{
  \hrule height 4pt
  \vskip 0.25in
  \vskip -\parskip%
}
\newcommand{\bottomtitlebar}{
  \vskip 0.29in
  \vskip -\parskip
  \hrule height 1pt
  \vskip 0.09in%
}
\newcommand{\appendixtitle}[1]{{\phantomsection\hsize\textwidth\linewidth\hsize %
  \vskip 0.1in \toptitlebar \centering{\Large\bf #1\par}\bottomtitlebar%
  \addcontentsline{toc}{section}{#1}}}

\begin{document}
\pagestyle{headings}
\mainmatter

\title{Pioneer Networks: Progressively Growing \\ Generative Autoencoder}
\titlerunning{Pioneer Networks}
\authorrunning{Ari Heljakka, Arno Solin, Juho Kannala}

\author{Ari Heljakka\inst{1,2} \and Arno Solin\inst{1} \and Juho Kannala\inst{1} \\% 
        \texttt{\{ari.heljakka, arno.solin, juho.kannala\}@aalto.fi}}
\institute{\!\!Aalto University \and \!\!GenMind Ltd} 

\maketitle

\begin{abstract}
  We introduce a novel generative autoencoder network model that learns to encode and reconstruct images with high quality and resolution, and supports smooth random sampling from the latent space of the encoder. Generative adversarial networks (GANs) are known for their ability to simulate random high-quality images, but they cannot reconstruct existing images. Previous works have attempted to extend GANs to support such inference but, so far, have not delivered satisfactory high-quality results. Instead, we propose the Progressively Growing Generative Autoencoder (PIONEER) network which achieves high-quality reconstruction with $128{\times}128$ images without requiring a GAN discriminator. We merge recent techniques for progressively building up the parts of the network with the recently introduced adversarial encoder--generator network. The ability to reconstruct input images is crucial in many real-world applications, and allows for precise intelligent manipulation of existing images. We show promising results in image synthesis and inference, with state-of-the-art results in CelebA inference tasks.
\end{abstract}

\section{Introduction}
Recent progress in generative image modelling and synthesis using generative adversarial networks (GANs, \cite{goodfellow2014}) has taken us closer to robust high-quality image generation. In particular, progressively growing GANs (ProgGAN, \cite{karras2017}) can synthesize realistic high-resolution images with unprecedented quality. For example, given a training dataset of real face images, the models learnt by ProgGAN are capable of synthesizing face images that are visually indistinguishable from face images of real people.

However, GANs have no inference capability. While useful for understanding representations and generating content for training other models, the capability for realistic image synthesis alone is not sufficient for most applications. Indeed, in most computer vision tasks, the learnt models are used for feature extraction from existing real images. This motivates generative autoencoder models that allow both generation and reconstruction so that the mapping between the latent feature space and image space is bi-directional. For example, image enhancement and editing would benefit from generation and inference capabilities \cite{brock2016}. In addition, unsupervised learning of generative autoencoder models would be widely useful in semi-supervised recognition tasks. Yet, typically the models such as variational autoencoders (VAEs, \cite{kingma2014,rezende2014}) generate samples not as realistic nor rich with fine details as those generated by GANs. Thus, there have been many efforts to combine GANs with autoencoder models \cite{larsen2015,makhzani2015,brock2016,Donahue2017,dumoulin2016D,ulyanov2017,Rosca2017}, but none of them has reached results comparable to ProgGAN in quality.

\begin{figure}[!t]
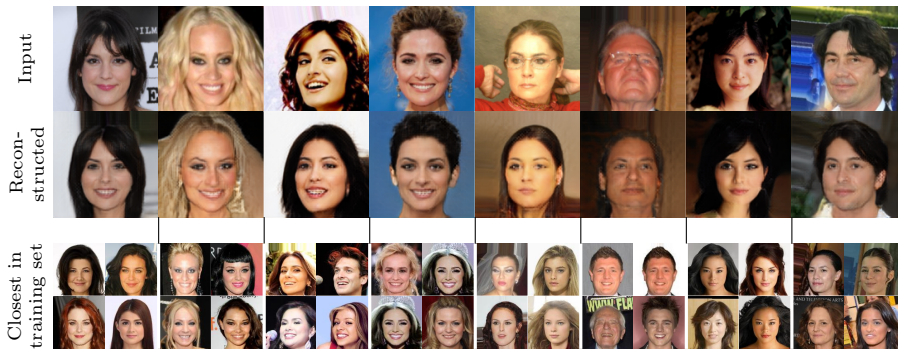

  \centering\scriptsize
  \setlength{\figurewidth}{0.115\textwidth}
  \setlength{\figureheight}{\figurewidth}
  \begin{tikzpicture}
    \foreach \i in {0,...,15}
      \node[draw=white,fill=black!20,minimum size=\figurewidth,inner sep=0pt]
        (\i) at ({\figurewidth*int((\i)/2)},{-\figureheight*mod(\i,2)})
        {\includegraphics[width=\figurewidth]{./fig/fig1/reconstruction_\i.png}};

    \foreach \i in {0,...,31}
      \node[draw=white,fill=black!20,minimum size=0.5\figurewidth,inner sep=0pt]
        (\i) at ({\figurewidth/2*int((\i)/2)-\figurewidth/4},{-\figurewidth/2*mod((\i,2))-2*\figurewidth}) {\includegraphics[width=0.5\figurewidth]{./fig/fig1/closest/sample-\i.jpg}};

    \foreach \i in {0,...,6}
      \draw[very thick,white] 
        ({\i*\figurewidth+\figurewidth/2},{-1.75*\figurewidth}) -- 
        ({\i*\figurewidth+\figurewidth/2},{-2.75*\figurewidth});
    \foreach \i in {0,...,6}
      \draw[thin,black] 
        ({\i*\figurewidth+\figurewidth/2},{-1.5*\figurewidth}) -- 
        ({\i*\figurewidth+\figurewidth/2},{-1.75*\figurewidth});

    \node[rotate=90,text width=\figurewidth, align=center] at (-0.75\figurewidth,-0*\figureheight) {Input};
    \node[rotate=90,text width=\figurewidth, align=center] at (-0.75\figurewidth,-1*\figureheight) {Recon\-structed};
    \node[rotate=90,text width=\figurewidth, align=center] at (-0.75\figurewidth,-2.25*\figureheight) {Closest in training set};              
              
  \end{tikzpicture}
  \caption{Examples of \PIONEER network reconstruction quality in $128{\times}128$ resolution (randomly chosen images from the \CelebA test set). Here, images are encoded into 512-dimensional latent feature vector and simply decoded back to the original dimensionality. Below each image pair, we show the four closest face matches to the input image in the training set (with respect to structural similarity \cite{Zhou+Bovik+Sheikh+Simoncelli:2004} of the face, cropped as in \cite{karras2017}).}
  \label{fig:reconstructions}
  \vspace*{-1em}
\end{figure}

In this paper, we propose the {\bf P}rogress{\bf I}vely gr{\bf O}wi{\bf N}g g{\bf E}nerative auto{\bf E}ncode{\bf R} (\PIONEER) network that extends the principle of progressive growing from purely generative GAN models to autoencoder models that allow both generation and inference. That is, we introduce a novel generative autoencoder network model that learns to encode and reconstruct images with high quality and resolution as well as to produce new high-quality random samples from the smooth latent space of the encoder. Our approach formulates its loss objective following \cite{ulyanov2017}, and we utilize spectral normalization \cite{miyato2018} to stabilize training---to gain the same effect as the `improved' Wasserstein loss \cite{gulrajani2017} used in~\cite{karras2017}. \sloppy

Similarly to \cite{ulyanov2017}, our approach contains only two networks, an encoder and a generator. The encoder learns a mapping from the image space to the latent space, while the generator learns the reciprocal mapping. Examples of reconstructions obtained by mapping a real input face image to the latent space and back using our learnt encoder and generator networks at $128{\times}128$ resolution are shown in Figure~\ref{fig:reconstructions}. Examples of synthetic face images generated from randomly sampled latent features by the generator are shown in Figure~\ref{fig:samples}. In these examples, the model is trained using the \CelebA \cite{Liu:2015} and \CelebAHQ \cite{karras2017} datasets in a completely unsupervised manner. We also demonstrate very smooth interpolation between tuples of test images that the network has never seen before, a task that is difficult and tedious to carry out with GANs.

In summary, the key contributions and results of this paper are:{\it (i)}~We propose a generative image autoencoder model whose architecture is built up progressively, with a balanced combination of reconstruction and adversarial losses, but without a separate GAN-like discriminator; {\it (ii)}~We show that at least up to $128{\times}128$ resolution, this model can carry out inference on input images with sharp output, and up to $256{\times}256$ resolution, it can generate sharp images, while having a simpler architecture than the state-of-the-art of purely generative models; {\it (iii)}~Our model gives improved image reconstruction results with larger image resolutions than previous state-of-the-art on \CelebA. The PyTorch source code of our implementation is available at \url{https://aaltovision.github.io/pioneer}.

\section{Related Work}
\label{sec:related-work}
\PIONEER networks belong to the family of generative models, with variational autoencoders (VAEs), autoregressive models, GAN variants, and other GAN-like models (such as \cite{li2015}). The core idea of a GAN is to jointly train so-called generator and discriminator networks so that the generator learns to output samples from the same distribution as the training set \cite{goodfellow2014}, when given random input vectors from a low-dimensional latent space, and the discriminator simultaneously learns to distinguish between the synthetic and real training samples. The generator and discriminator are differentiable, jointly learnt via backpropagation using alternating optimization of an adversarial loss, where the discriminator is updated to maximize the probability of correctly classifying real and synthetic samples and the generator is updated to maximize the probability of discriminator making a mistake. Upon convergence, the generator learns to produce samples that are indistinguishable from the training samples (within the limits of the discriminator network's capacity).

Making the aforementioned training process stable has been a challenge, but the Wasserstein GAN \cite{Arjovsky2017b} improved the situation by adopting a smooth metric for the distance between the two probability distributions \cite{gulrajani2017}. In Karras~\etal \cite{karras2017}, the Wasserstein GAN loss from \cite{gulrajani2017} is combined with the idea of progressively growing the layers and image resolution of the generator and discriminator during training, yielding excellent image synthesis results. Progressive growing has been used successfully also, for example, by \cite{zhang2016}. There is also a line of work on other regularizers that stabilize the training (\eg \cite{RothLNH17,miyato2018,qi2017}).

However, it is well understood that the capability for realistic image synthesis alone is not sufficient for applications and there is a need for better unsupervised feature learning methods that are able to capture the semantically relevant dependencies of input data into a compact latent representation \cite{Donahue2017}. In their basic form, GANs are not suitable for this purpose as they do not provide means of learning the inverse mapping that projects the data back to latent space.

Nevertheless, there have been many recent efforts which utilize adversarial training for learning bi-directional generative models that would allow both image synthesis and reconstruction in a manner similar to autoencoders. For example, the recent works \cite{Donahue2017} and \cite{dumoulin2016D} simultaneously proposed an approach that employs three deep neural networks (generator, encoder, and discriminator) for learning bi-directional mappings between the data space and latent space. Instead of just samples, the discriminator is trained to discriminate tuples of samples with their latent codes, and it is shown that at the global optimum the generator and encoder learn to invert each other. Further, several others have proposed 3-network approaches that add some form of reconstruction loss and combine ideas of GAN and VAE: \cite{larsen2015} extends VAE with a GAN-like discriminator for sample space (also used by \cite{brock2016}), \cite{makhzani2015,mescheder2017} do the same with a GAN-like discriminator for the latent space, and \cite{Rosca2017} adds yet another discriminator (for the VAE likelihood term). While the previous methods have advanced the field, they still have not been able to simultaneously provide high quality results for both synthesis and reconstruction of high resolution images. Most of these methods struggle with even $64{\times}64$ images.

Recently, Ulyanov~\etal \cite{ulyanov2017} presented an autoencoder architecture that simply consists of two deep networks, a generator $\theta$ and encoder $\phi$, representing mappings between the latent space and data space, and trained with a combination of adversarial optimization and reconstruction losses. That is, given the data distribution $X$ and a simple prior distribution $Z$ in the latent space, the updates for the generator aim to minimize the divergence between $Z$ and $\vphi(\vtheta(Z))$, whereas the updates for the encoder aim to minimize the divergence between $Z$ and $\vphi(X)$ and simultaneously maximize the divergence between $\vphi(\vtheta(Z))$ and $\vphi(X)$. In addition, the adversarial loss is supplemented with reconstruction losses both in the latent space and image space to ensure that the mappings are reciprocal (\ie\ $\vphi(\vtheta(\vz)) \simeq \vz$ and $\vtheta(\vphi(\vx)) \simeq \vx$). The results of \cite{ulyanov2017} are promising regarding both synthesis and reconstruction but the images still have low resolution. Scaling to higher resolutions requires a larger network which makes adversarial training less stable.

We combine the idea of progressive network growing \cite{karras2017} with the adversarial generator--encoder (AGE) networks of \cite{ulyanov2017}. However, the combination is not straightforward, and we needed to identify a proper set of techniques to stabilize the training. In summary, our contributions result in a model that is simpler than many previous ones (\eg~having a large discriminator network just for the purpose of training the generator is wasteful and can be avoided), provides better results than \cite{ulyanov2017} already in small ($64{\times}64$) resolutions, and enables training and good results with larger image resolutions than previously possible. The differences to \cite{Rosca2017}, \cite{Donahue2017}, and, for example, \cite{tabor2018} are substantial enough to perceive by quick visual comparison.

\section{\PIONEER Networks}
\label{sec:pioneer-networks}
Our generative model achieves three key goals that define a good encoder--decoder model: {\it (i)}~faithful reconstruction of the input sample, {\it (ii)}~high sample quality (whether random samples or reconstructions), and {\it (iii)}~rich representations. The final item can be reformulated as a `well-behaved' latent space that lends itself to high-quality interpolations between given test samples and captures the diversity of features present in the training set. Critically, these requirements are strictly parametrized by our target resolution---there are several models that achieve many of the said goals up to $32{\times}32$ image resolution, but very few that have shown good results beyond $64{\times}64$ resolution.

\PIONEER networks achieve the reconstruction and representation goals up to $128{\times}128$ resolution and the random sample generation up to $256{\times}256$ resolution, while using a combination of simple principles. A conceptual description in the next subsection is  followed by some theory (Sec.~\ref{sec:encode-decoder}) and more practical implementation details (Sec.~\ref{sec:model}).

\begin{figure}[!t]
\centering
\resizebox{.4\textwidth}{!}{%
\begin{tikzpicture}\normalsize

  \draw [line width=10pt,rotate=9,line cap=round,<-,draw=green!30, postaction={decorate},decoration={text along path, raise=-4ex, reverse path,
text={|\large\color{black}|Encoder optimization loop},text align=center}] (0,-4.25) arc (-90:250:4.25cm);

  \draw [line width=10pt,rotate=188,line cap=round,<-,draw=blue!50,postaction={decorate},decoration={text along path, raise=-4ex,
text={|\large\color{black}|Decoder optimization loop},text align=center}] (0,-5.0) arc (-90:250:5.0cm);

  \node[draw=white,minimum size=1.2cm, inner sep=0] (A) at (-1.7,-2) {\includegraphics[width=12mm]{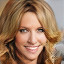}};
  \node[draw=white,minimum size=1.2cm, inner sep=0] (B) at ( 1.7,-2) {\includegraphics[width=12mm]{./fig/fig2/A-4}};

  \tikzstyle{block} = [draw=black,fill=black!20,minimum size=2.5mm]
  \foreach \i in {0,...,7} \node[block] at (0.125-1.7+\i/4-1,-0.75) {};
  \foreach \i in {0,...,7} \node[block] at (0.125+1.7+\i/4-1,-0.75) {};
  \foreach \i in {0,...,5} \node[block] at (0.125-1.7+\i/4-0.75,0.0) {};
  \foreach \i in {0,...,5} \node[block] at (0.125+1.7+\i/4-0.75,0.0) {};
  \foreach \i in {0,...,3} \node[block] at (0.125-1.7+\i/4-0.5,0.75) {};
  \foreach \i in {0,...,3} \node[block] at (0.125+1.7+\i/4-0.5,0.75) {};
  \foreach \i in {0,...,11} \node[block,minimum size=3.333mm] at (\i/3+0.16665-2,1.75) {};

  \node[rotate=90] at (-3.3,0) {encoder};
  \node[rotate=-90] at (3.3,0) {decoder/generator};
  \node[] at (0,2.7) {\huge $\vect{z}$};
  \node[] at (0,-4.15) {\huge $\Delta\vect{x}$};
  \node[] at (0,5.15) {\huge $\Delta\vect{z}$};
  \node[] at (-1.7,-3) {Input};
  \node[] at (1.7,-3) {Output};

  \draw[-latex] (A) -> (-1.7,-0.9);
  \draw[-latex] (-1.7,-0.6) -> (-1.7,-0.2);
  \draw[-latex] (-1.7, 0.2) -> (-1.7,0.55);
  \draw[-latex] (-1.7, 0.95) -> (-1.7,1.5);

  \draw[-latex] (1.7,-0.9) -> (B);
  \draw[-latex] (1.7,-0.2) -> (1.7,-0.6);
  \draw[-latex] (1.7,0.55) ->(1.7, 0.2);
  \draw[-latex] (1.7,1.5) -> (1.7, 0.95);
      
\end{tikzpicture}}
\hspace*{\fill}
\resizebox{.58\textwidth}{!}{%
\begin{tikzpicture}\tiny
  \newcommand{\pow}[1]{\pgfmathparse{int(2^(#1+2))}\pgfmathresult}

  \foreach \i in {1,...,4} 
    \node[] (A-\i) at (2.1*\i+1-2.1,4.25) {\includegraphics[width=8mm]{./fig/fig2/A-\i}};
  \foreach \i in {1,...,4} 
    \node[] (B-\i) at (2.1*\i+1-2.1,-0.25) {\includegraphics[width=8mm]{./fig/fig2/A-\i}};

  \draw[-latex,black] (0,-.75) -> (8, -.75); 
  \draw[dashed,thin] (0,2) -- (8,2);
  \foreach \i in {1,...,4}
    \draw[-latex,thin] (A-\i) -> (2.1*\i-1.1,2.1);
  \foreach \i in {1,...,4}
    \draw[-latex,thin] (2.1*\i-1.1,2) -> (B-\i);

  \tikzstyle{block} = [draw=black,fill=black!20,minimum size=0.25mm,thin]
  \foreach \j in {0,...,3} 
    \foreach \i in {0,...,6} \node[block] at (2.1*\j+\i/6+1-0.5,2) {};

  \tikzstyle{block} = [minimum width=1cm,minimum height=2mm,inner sep=0,draw=black,fill=white]
  \foreach \i in {1,...,4} 
    \foreach \j in {0,...,\i} 
      \node[block,minimum width={0.75cm+0.25cm*\j}] (b-\i-\j) at (2.1*\i+1-2.1,1.5-.22*\j) {$\pow{\j}{\times}\pow{\j}$};

  \foreach \i in {1,...,4} 
    \foreach \j in {0,...,\i} 
      \node[block,minimum width={0.75cm+0.25cm*\j}] (t-\i-\j) at (2.1*\i+1-2.1,2.5+.22*\j) {$\pow{\j}{\times}\pow{\j}$};

  \node[] at (6,-.9) {Training progresses};
  \node[rotate=90] at (0.1,3.00) {\scriptsize Encoder, $\phi$};
  \node[rotate=90] at (0.1,1.00) {\scriptsize Decoder, $\theta$};
  \node[rotate=90] at (0.1,4.25) {\scriptsize In};  
  \node[rotate=90] at (0.1,-0.25) {\scriptsize Out};    
\end{tikzpicture}}
\caption{The network grows in phases during which the image resolution doubles. The adversarial/reconstructive training criterion is continuously applied at each step, adapting to the present input--output resolution. The circular arrows illustrate the two modes of learning: {\it (i)}~reconstruction of real training samples, and {\it (ii)}~reconstruction of the latent representation of randomly generated samples.}
\label{fig:sketch}
\vspace*{-1em}
\end{figure}

\subsection{Intuition}
\label{sec:intuition}
The defining training and architecture principles of \PIONEER networks are shown in Figure~\ref{fig:sketch}; on the left hand side, the competing objectives are presented in the double loop, and on the right, the progressively growing structure of the network is shown stepping up through $4{\times}4, 8{\times}8, 16{\times}16, \ldots$, doubling the resolution in each phase. The input $\vx$ is squeezed through the encoder into a latent representation $\vz$, which on the other hand is again decoded back to an image $\hat{\vx}$. The motivation behind the progressively growing setup is to encourage the network to catch the fundamental structure and variation in the inputs at lower resolutions to help the additional layers specialize in fine-tuning and adding details and nuances when reaching the higher resolutions.

The network has encoder--decoder structure with no {\it ad~hoc} components (such as separate discriminators as in \cite{brock2016,larsen2015,Rosca2017,makhzani2015,mescheder2017}). Similar to GANs, the encoder and decoder are not trained as one, but instead as if they were two competing networks. This requires the encoder to become sensitive to the difference between training samples and generated (decoded) samples, and the decoder to keep making the difference smaller and smaller. While GANs achieve this with the complexity cost of a separate discriminator network, we choose to just learn to encode the samples in a source-dependent manner. This encoding could be, then, followed by a classification layer, but instead we train the encoder so that the distribution of latent codes of training samples \emph{approach} a certain reference distribution, while the distribution of codes of generated samples \emph{diverges} from it (see AGE \cite{ulyanov2017}).

\subsection{Encoder--Decoder Losses}
\label{sec:encode-decoder}
As in variational autoencoders, we choose the Kullback--Leibler (KL) divergence as the metric in latent space. Our reference distribution is unit Gaussian with a diagonal covariance matrix. Each sample $\vx \in X$ is encoded into a latent vector $\vz \in Z$, giving rise to the posterior distribution $q_\phi(\vz\mid\vx)$ on a $d$-dimensional sphere. The KL-divergence between such a distribution and a $d$-dimensional unit Gaussian is (see the reasoning in \cite{ulyanov2017}, but with the following corrections):
\begin{align}  
  \mathrm{KL}[q_\phi(\vz\mid\vx) \,\|\, \mathcal{N}(\vect{0},\MI)] &= -\frac{d}{2} + \sum_{j=1}^{d}\bigg[\frac{\sigma_j^2 + \mu_j^2}{2} - \log(\sigma_j)\bigg],
\end{align}
where $\mu_j$ and $\sigma_j$ are the empirical sample mean and standard deviation of the encoded samples in the latent vector space with respect to dimension $j=1,2,\ldots,d$, and $\mathcal{N}(\vect{0},\MI)$ denotes the unit Gaussian.

The encoder $\phi$ and decoder $\theta$ are connected via two reconstruction error terms. We measure reconstruction error $L_\mathcal{X}$ with L1 distance in sample space $\mathcal{X}$ for the encoder, and code reconstruction error $L_\mathcal{Z}$ with cosine distance in latent code space $\mathcal{Z}$ for the decoder, as follows:
\begin{align}
  L_\mathcal{X}(\vtheta,\vphi) &= \mathbb{E}_{\vect{x} \sim X} \| \vect{x} -\vtheta(\vphi(\vect{x})) \|_1, \\
  L_\mathcal{Z}(\vtheta,\vphi) &= \mathbb{E}_{\vect{z} \sim Z}{[1 - \vect{z}\T \vphi(\vtheta(\vect{z}))]},
\end{align}
where $X$ are the training samples and $Z$ random latent vectors, with $\vect{z}$ and $\vphi(\vect{x})$ normalized to unity.

In other words, a training sample is encoded into the latent space and then decoded back into a generated sample. A random latent vector is decoded into a random generated sample that is then fed back to the encoder (Fig.~\ref{fig:sketch}). This provides an elegant solution to forcing the network to learn to reconstruct training images.
The total loss function of the encoder $L_\phi$ and decoder $L_\theta$ are, then:
\begin{align}
  L_\phi &= \phantom{-}\mathrm{KL}[q_\phi(\vz\mid\vx) \,\|\, \mathcal{N}(\vect{0},\MI)] - \mathrm{KL}[q_\phi(\vz\mid\hat{\vx}) \,\|\, \mathcal{N}(\vect{0},\MI)] + \lambda_\mathcal{X} L_\mathcal{X}, \label{eq:loss-phi} \\
  L_\theta &= -\mathrm{KL}[q_\phi(\vz\mid\vx) \,\|\, \mathcal{N}(\vect{0},\MI)] + \mathrm{KL}[q_\phi(\vz\mid\hat{\vx}) \,\|\, \mathcal{N}(\vect{0},\MI)] + \lambda_\mathcal{Z}L_\mathcal{Z}, \label{eq:loss-theta}
\end{align}
where $\vx \sim X$ and $\hat{\vx}$ = $\vtheta(\vz)$ with $\vz \sim \mathcal{N}(\vect{0},\MI)$. We fix the hyper-parameters $\lambda_\mathcal{X}$ and $\lambda_\mathcal{Z}$ so they can be read as scaling constants. In practical implementation, we can simplify the decoder loss to only account for
\begin{align}
  L_\theta &= \mathrm{KL}[q_\phi(\vz\mid\hat{\vx}) \,\|\, \mathcal{N}(\vect{0},\MI)] + \lambda_\mathcal{Z}L_\mathcal{Z}. \label{eq:loss-theta2}
\end{align}
The training is adversarial in the sense that we use each loss function in turn, first freezing the decoder weights and training only with the loss \eqref{eq:loss-phi}, and then freezing the encoder weights and training only with the loss \eqref{eq:loss-theta2}.

However, in Ulyanov~\etal\ \cite{ulyanov2017}, this approach was only shown to work with AGE on images up to $64{\times}64$ resolution. Beyond that, we need a larger network architecture, which is unlikely to work with AGE alone. We confirmed this by trying out a straightforward extension of AGE to $128{\times}128$ resolution (by visual examination and via results in Table~\ref{tbl:results}). In contrast, to stabilize training, our model will increase the size of the network progressively, following \cite{karras2017}, and utilize the following techniques.

\subsection{Model and Training}
\label{sec:model}
The training uses a convolution--deconvolution architecture typically used in generative models, but here, the model is built up progressively during training, as in \cite{karras2017}. We start training on low resolution images ($4{\times}4$), bypassing most of the network layers. We train each intermediate phase with the same number of samples. In the first half of each consequtive phase, we start by adding a trivial downsampling (encoder) and upsampling (decoder) layer, which we gradually replace by fading in the next convolutional--deconvolutional layers simultaneously in the encoder and the decoder, in lockstep with the input resolution which is also faded in gradually from the previous to the new doubled resolution ($8{\times}8$ etc.). During the second half of each phase, the architecture remains unchanged. After the first half of the target resolution phase, we no longer change the architecture.

We train the encoder and the generator with loss \eqref{eq:loss-phi} and \eqref{eq:loss-theta2} in turn, utilizing various stabilizing factors as follows. The architecture of the convolutional layers in \PIONEER networks largely follows yet simplifies the symmetric structure in ProgGAN (see Table 2 of \cite{karras2017}), with the provision of replacing its discriminator with an encoder. This requires removing the binary classifier, allowing us to connect the encoder and decoder directly via the 512-dimensional latent vector. We also remove the minibatch standard deviation layer, as it is sensitive to batch-level statistics useful for a GAN discriminator but not for an encoder.

For stabilizing the training, we employ equalized learning rate and pixelwise feature vector normalization in the generator \cite{karras2017}, buffer of images created by previous generators \cite{shrivastava2016}, and encoder spectral normalization \cite{miyato2018}. We use ADAM \cite{adam} with $\beta_1 = 0, \beta_2 = 0.99, \epsilon=10^{-8}$ and learning rate 0.001. We use 2 generator updates per 1 encoder update. For result visualization (but not training), we use an exponential running average for the weights of the generator over training steps as in \cite{karras2017}. Of these techniques, spectral normalization warrants some elaboration.

To stabilize the training of generative models, it is important to consider the function space within which the discriminator must fit in general, and, specifically, controlling its Lipschitz constant. ProgGAN uses improved Wasserstein loss \cite{gulrajani2017} to keep the Lipschitz constant close to unity. However, this loss formulation is not immediately applicable to the slightly more complex AGE-style loss formulation, so instead, we adopted GAN spectral normalization \cite{miyato2018} to serve the same purpose. In this spectral normalization approach, the spectral norm of each layer of the encoder (discriminator) network is constrained directly at each computation pass, allowing the network to keep the Lipschitz constant under control. Crucially, spectral normalization does not regularize the network via learnable parameters, but affects the scaling of network weights in a data-dependent manner.

In our experiments, it was evident that without such a stabilizing factor, the progressive training would not remain stable beyond $64{\times}64$ resolution. Spectral normalization solved this problem unambiguously: without it, the training of the network was consistently failing, while with it, the training almost consistently converged. Other strong stabilization methods, such as the penalty on the weighted gradient norm \cite{RothLNH17}, might have worked here as well.

\begin{figure}[!t]
\centering\scriptsize
\setlength{\figurewidth}{.0475\textwidth}
\setlength{\figureheight}{\figurewidth}
\begin{tikzpicture}

  \foreach \i in {1,...,40}
    \node[draw=white,fill=black!20,minimum size=\figurewidth,inner sep=0pt]
      (\i) at ({\figurewidth*mod(\i-1,20)},{-\figureheight*int((\i-1)/20)})
      {\includegraphics[width=\figurewidth]{./fig/fig3/64/093750_\i.jpg}};

  \setlength{\figurewidth}{2\figurewidth}
  \setlength{\figureheight}{\figurewidth}  
  \foreach \i in {1,...,20}
    \node[draw=white,fill=black!20,minimum size=\figurewidth,inner sep=0pt]
      (\i) at ({0.25\figurewidth+\figurewidth*mod(\i-1,10)},{-1.25\figureheight-\figureheight*int((\i-1)/10)})
      {\includegraphics[width=\figurewidth]{./fig/fig3/128/037500_\i.jpg}};

  \setlength{\figurewidth}{2\figurewidth}
  \setlength{\figureheight}{\figurewidth}

  \node[draw=white,fill=black!20,minimum size=\figurewidth,inner sep=0pt]
    at ({0.375\figurewidth+\figurewidth*0},{-1.875\figureheight})
       {\includegraphics[width=\figurewidth]{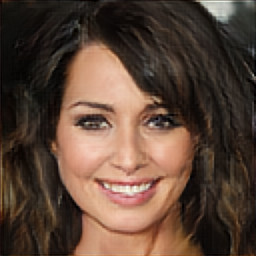}};
  \node[draw=white,fill=black!20,minimum size=\figurewidth,inner sep=0pt]
    at ({0.375\figurewidth+\figurewidth*1},{-1.875\figureheight})
       {\includegraphics[width=\figurewidth]{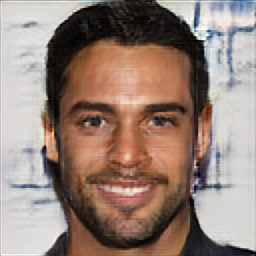}};
  \node[draw=white,fill=black!20,minimum size=\figurewidth,inner sep=0pt]
    at ({0.375\figurewidth+\figurewidth*2},{-1.875\figureheight})
       {\includegraphics[width=\figurewidth]{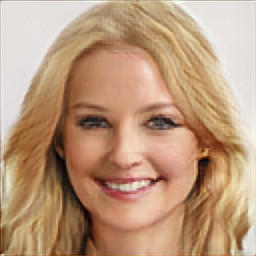}};
  \node[draw=white,fill=black!20,minimum size=\figurewidth,inner sep=0pt]
    at ({0.375\figurewidth+\figurewidth*3},{-1.875\figureheight})
       {\includegraphics[width=\figurewidth]{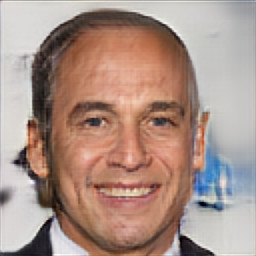}};
  \node[draw=white,fill=black!20,minimum size=\figurewidth,inner sep=0pt]
    at ({0.375\figurewidth+\figurewidth*4},{-1.875\figureheight})
       {\includegraphics[width=\figurewidth]{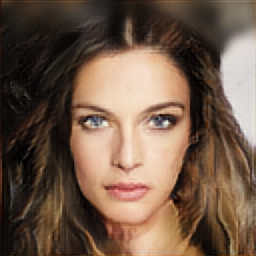}};

\end{tikzpicture}
\caption{Randomly generated face image samples with \PIONEER networks using \CelebA for training at resolutions $64{\times}64$ (top) and $128{\times}128$ (middle), and using \CelebAHQ for $256{\times}256$ (bottom).}
\label{fig:samples}
\vspace*{-1em}
\end{figure}

\section{Experiments}
\label{sec:experiments}
\PIONEER networks are more most immediately applicable to learning image datasets with non-trivial resolutions, such as \CelebA \cite{Liu:2015}, LSUN \cite{Yu:2015}, and \textsc{ImageNet}. Here, we run experiments on \CelebA and \CelebAHQ \cite{karras2017} (with training/testing split 27000/3000) and LSUN bedrooms. For comparing with previous works, we also include \Cifar, although its low-resolution images ($32{\times}32$) were not expected to be most relevant for the present work.

Training with high resolutions is relatively slow in both ProgGAN and our method, but we believe that significant speed optimization is possible in future work. In fact, it is noteworthy that you \emph{can} train these models for a long time without running into typical GAN problems, such as `mode collapse' or ending up oscillating around a clearly suboptimal point. We trained the \PIONEER model on \CelebA with one Titan~V GPU for 5~days up to $64{\times}64$ resolution (172 epochs), and another 8~days for $128{\times}128$ resolution. We separately trained on \CelebAHQ up to $256{\times}256$ resolution with four Tesla~P100 GPUs for 10~days (1600 epochs), and on LSUN with two Tesla~P100 GPUs for 9~days.

Throughout the training, we kept the hyper-parameters fixed at $\lambda_\mathcal{Z} = 1000\,d$ and $\lambda_\mathcal{X} = 10\,d$, where $d$ is the dimensionality of the latent space (512), taking advantage of the hyper-parameter search done by \cite{ulyanov2017}. After the progressive growth phase of the training, we switched to $\lambda_\mathcal{X} = 15\, d$ to emphasize sample reconstruction \cite{ulyanovGithub}.

\subsection{\CelebA and \CelebAHQ}
\label{sec:CelebA}
The \CelebA dataset \cite{Liu:2015} contains over 200k images with various resolutions that can be square-cropped to $128{\times}128$. \CelebAHQ \cite{karras2017} is a subset of 30k of those images that have been improved and upscaled to $1024{\times}1024$ resolution. We train with \CelebA up to $128{\times}128$ resolution, and with \CelebAHQ up to $256{\times}256$. In order to compare with previous works, we also trained our network for $64{\times}64$ images from \CelebA.

We ran our experiments as follows. Following the approach described in Section~\ref{sec:model}, we trained the network progressively through each intermediate resolution until we reach the target resolution ($64{\times}64$, $128{\times}128$, or $256{\times}256$), for the same number of steps in each stage. For the final stage with the target resolution, we would continue training for as long as the Fr\'echet Inception Distance (FID, \cite{heusel2017}) measures of the randomly generated samples showed improvements. During the progression of the input resolution, we adapted minibatch size to accommodate for the available memory.

For random sampling metrics, we use FID and Sliced Wasserstein Distance (SWD, \cite{rabin2015}) between the training distribution and the generated distribution. FID measures the sample quality and diversity, while SWD measures the similarity in terms of Wasserstein distance (earth mover's distance). Batch size is 10000 for FID and 16384 for SWD. For reconstruction metrics, we use the root-mean-square error (RMSE) between the original and the reconstructed image.

We present our results in three ways. First, the model must be able to reconstruct random test set images and retain both sufficient quality and faithfulness of the reconstruction. Often, there is a trade-off between the two \cite{rosca2018}. Previous models have often seemed to excel with respect to the quality of the reconstruction image, but in fact, the reconstruction turns out to be very different from the original (such as a different person's face). Second, we must be able to randomly sample images from the latent space of the model, and achieve sufficient quality and diversity of the images. Third, due to its inference capability, \PIONEER networks can show interpolated results between input images without any additional tricks, such as first solving a regression optimization problem for the image, as often done with GANs (\eg~\cite{radford2015}).

\begin{table}[!tb]
  \caption{Comparison of Fr\'echet Inception Distance (FID) against 10,000 training samples, Sliced Wasserstein Distance (SWD) against 16384 samples, and root-mean-square error (RMSE) on test set, in the $64{\times}64$ and $128{\times}128$ \CelebA dataset on inference-capable networks. ProgGAN with L1 regression has the best overall sample quality (in FID/SWD), but not best reconstruction capability (in RMSE). A pretrained model by the author of \cite{ulyanov2017} was used for AGE for $64{\times}64$. For $128{\times}128$, we enlarged the AGE network to account for the larger inputs and a 512-dimensional latent vector, trained until the training became unstable. ALI was trained on default \CelebA settings following \cite{dumoulin2016D} for 123 epochs. The error indicates one standard deviation for separate sampling batches of a single (best) trained model. For all numbers, \textbf{smaller is better}.}
  \label{tbl:results}
  \centering
  {\vspace*{3pt}\footnotesize\noindent\scriptsize%
  \begin{tabular*}{\textwidth}{@{\extracolsep{\fill}} lcccccc}
  \toprule
  & \multicolumn{3}{c}{$64{\times}64$} & \multicolumn{3}{c}{$128{\times}128$} \\
  \cmidrule(lr){2-4} \cmidrule(lr){5-7}
  & FID & SWD & RMSE & FID & SWD & RMSE \\  
  \midrule
  ALI  & $58.88 \pm 0.19$ & $25.58 \pm 0.35$ & $18.00 \pm 0.21$ & --- & --- & --- \\
  AGE  & $26.53  \pm 0.08$ & $17.87  \pm 0.11$ & $4.97  \pm 0.06$ & $154.79  \pm 0.43$ & $22.33  \pm 0.74$ & $9.09  \pm 0.07$ \\
  ProgGAN/L1 & {{$\bf7.98 \bf\pm \bf0.06$}} & {$\bf 3.54 \bf\pm \bf0.40$} & $2.78  \pm 0.05$ & --- & --- & --- \\  
  \PIONEER &   {$8.09  \pm 0.05$} & $5.18  \pm 0.19$ & {$\bf 1.82 \pm 0.02$} & {$\bf 23.15 \pm 0.15$} & {$\bf 10.99 \pm 0.44$} & {$\bf 8.24 \pm 0.15$} \\
  \bottomrule
  \end{tabular*}}

  \vspace*{-1em}
\end{table}

\vspace*{-6pt}
\subsubsection{Reconstruction.} Given an unseen test image, the model should be able to encode the relevant information (such as hair color, skin color, facial expression, \etc) and decode it into a natural-looking face image expressing the features. As this is not image compression, the model does not aim to replicate the input image {\em per~se}, but capture the essentials. In Figure~\ref{fig:reconstructions}, we show reconstruction examples for random test images in \CelebA (at $128{\times}128$ resolution) with \PIONEER. Under the reconstruction images we show the four closest samples in the training set (with respect to structural similarity \cite{Zhou+Bovik+Sheikh+Simoncelli:2004} of the face as cropped in \cite{karras2017}).

We compare reconstructions against inference-capable models: AGE \cite{ulyanov2017} and ALI \cite{dumoulin2016D}. We also train ProgGAN for reconstruction as follows (similar attempts have been done in, \eg, \cite{radford2015,lipton2017,luo2017,creswell2016}). We train the network normally until convergence, and then simply connect the latent vector of the discriminator to serve also as the latent input for the generator (properly normalized). Finally, we re-train the discriminator--generator network as an autoencoder that attempts to reconstruct input images, with L1 reconstruction loss. During this training, we only modify the discriminator subnetwork, since allowing the generator to change would inevitably lead to lower-quality generated images. (We also attempted training another fully connected layer on top of the existing hidden layer, but did not see improved results, and training became almost prohibitively slow.) Like most of the previous results, we find that the network (ProgGAN/L1) can fairly well reconstruct samples that it has generated itself, but performs much worse when given new real input images.

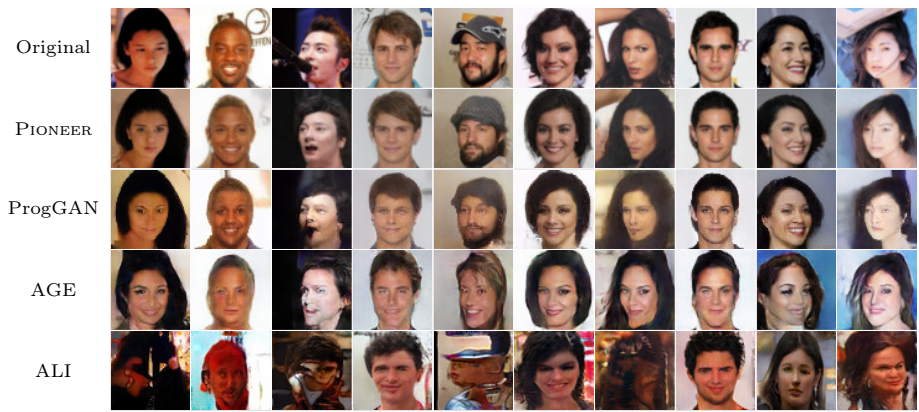
\begin{figure}[!t]
\centering\scriptsize
\setlength{\figurewidth}{.088\textwidth}
\setlength{\figureheight}{\figurewidth}
\begin{tikzpicture}

  \tikzstyle{fig} = [draw=white,fill=black!20,minimum size=\figurewidth,inner sep=0];
  \newcommand{\figg}[1]{\includegraphics[width=.97\figurewidth]{./fig/fig4a/#1.jpg}}

  \foreach \i in {0,...,9} {%
     \node[fig] at (\figurewidth*\i,-0*\figureheight) {\figg{orig_\i}};
     \node[fig] at (\figurewidth*\i,-1*\figureheight) {\figg{pine_\i}};
     \node[fig] at (\figurewidth*\i,-2*\figureheight) {\figg{proggan_\i}};
     \node[fig] at (\figurewidth*\i,-3*\figureheight) {\figg{age_\i}};
     \node[fig] at (\figurewidth*\i,-4*\figureheight) {\figg{ALI1_\i}};
  }

  \node[] at (-1.2\figurewidth,-0*\figureheight) {Original};
  \node[] at (-1.2\figurewidth,-1*\figureheight) {\PIONEER};
  \node[] at (-1.2\figurewidth,-2*\figureheight) {ProgGAN};
  \node[] at (-1.2\figurewidth,-3*\figureheight) {AGE};
  \node[] at (-1.2\figurewidth,-4*\figureheight) {ALI};

\end{tikzpicture}
\caption{Comparison of reconstruction quality between \PIONEER, ALI, and AGE in $64{\times}64$. The first row in each set shows examples from the test set of \CelebA (not cherry-picked). The reproduced images of \PIONEER are much closer to the original than those of AGE or ALI. Note the differences in handling of the 5th image from the left.  ALI and AGE were trained as in Table~\ref{tbl:results}. (For more examples, see Supplementary Material.)}
\label{fig:reco64}
\vspace*{-1em}
\end{figure}

For networks that support both inference and generation, we can feed input images and evaluate the output image. In Figure~\ref{fig:reco64}, we show the output of each network for the given random \CelebA test set images. As seen from the figure, at $64{\times}64$ resolution, \PIONEER outperforms the baseline networks in terms of the combined output quality and faithfulness of the reconstruction.
At $64{\times}64$ resolution, \PIONEER's FID score of $8.09$ in Table~\ref{tbl:results} outperforms AGE and ALI, the relevant inference baselines. ProgGAN/L1 outperforms the rest in sample quality (FID/SWD), but is worse in faithfulness (RMSE).

Without modifications, ALI and AGE networks have not thus far been shown to be able to deal with $128{\times}128$ resolution or higher. We managed to run AGE for $128{\times}128$ resolution by enlarging the AGE network to account for the larger inputs and a 512-dimensional latent vector, and trained until the training became unstable. For ALI, enlarging the network for $128{\times}128$ was not tried. ProgGAN has been shown to excel in sample generation for higher resolutions, but as explained earlier, it is not designed for reconstruction or inference. Therefore we have only run it for $64{\times}64$, which already showed this difference.

\vspace*{-1em}
\subsubsection{Dreaming up random samples.} A model that focuses on reconstruction is unlikely to match the quality of the models that only focus on random sample generation. Even though our focus is on excelling in the former category, we do not fall far behind the state-of-the-art provided by ProgGAN in generating new samples. Figure~\ref{fig:samples} shows samples generated by \PIONEER at $64{\times}64$, $128{\times}128$, and $256{\times}256$ resolutions. The ProgGAN SWD results in \cite{karras2017} were based on a more aggressive cropping of the dataset, so the values are not comparable. For AGE and ALI, the FID and SWD scores are clearly worse (see Table~\ref{tbl:results}) even at low resolutions, and the methods do not generalize well to higher resolutions.

\begin{figure}[!t]
\centering\scriptsize
\setlength{\figurewidth}{.08\textwidth}
\setlength{\figureheight}{\figurewidth}
\newcommand\x{x}
\begin{tikzpicture}
  \foreach \i in {0,...,7}
    \foreach \j in {0,...,7}
      \node[draw=white,fill=black!20,minimum size=\figurewidth,inner sep=0]
        (\i) at ({\figurewidth*\i},{-\figureheight*\j})
        {\includegraphics[width=\figurewidth]{./fig/fig6/interpolations_5_480003_1.0_\j\x\i.jpg}};

  \foreach \i in {0,...,7}
     \node at ({\figurewidth*\i},{0.7\figureheight}) {\bf\pgfmathparse{int(\i+1)}\pgfmathresult};

  \def\myarray{{"A","B","C","D","E","F","G","H","I"}}
  \foreach \i in {0,...,7}
     \node at ({-0.7\figureheight},{-\figurewidth*\i}) {\bf\pgfmathparse{\myarray[\i]}\pgfmathresult};

  \node[inner sep=0] at 
    ({-1.5*\figurewidth},{-0\figureheight}) 
    {\includegraphics[width=1\figurewidth]{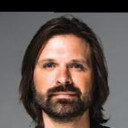}};
  \node[inner sep=0] at 
    ({-1.5*\figurewidth+10*\figurewidth},{-0\figureheight}) 
    {\includegraphics[width=1\figurewidth]{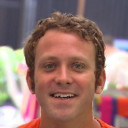}};        
  \node[inner sep=0] at 
    ({-1.5*\figurewidth},{-0\figureheight-7*\figureheight}) 
    {\includegraphics[width=1\figurewidth]{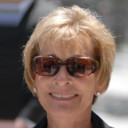}};
  \node[inner sep=0] at 
    ({-1.5*\figurewidth+10*\figurewidth},{-0\figureheight-7*\figureheight}) 
    {\includegraphics[width=1\figurewidth]{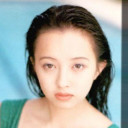}};

  \node[inner sep=0] at 
    ({-1.5*\figurewidth},{-0.75\figureheight}) {\bf In \#1};
  \node[inner sep=0] at 
    ({-1.5*\figurewidth+10*\figureheight},{-0.75\figureheight}) {\bf In \#2};
  \node[inner sep=0] at 
    ({-1.5*\figurewidth},{+0.75\figureheight-7*\figureheight}) {\bf In \#3};
  \node[inner sep=0] at 
    ({-1.5*\figurewidth+10*\figureheight},{+0.75\figureheight-7*\figureheight}) {\bf In \#4};
    
\end{tikzpicture}
\\[1em](Best viewed in high resolution / zoomed-in.)
\caption{Interpolation study on test set input images at $128{\times}128$ resolution. Unlike many works, we interpolate between the (reconstructions of) unseen test images given as input---not between images the network has generated on its own.} 
\label{fig:interpolation}
\vspace*{-1em}
\end{figure}

\vspace*{-1em}
\subsubsection{Inference capabilities.} Finally, we provide an example of input-based  interpolation between different (unseen) test images. In Figure~\ref{fig:interpolation} we have four different test images, one in each corner of the tile figure. Thus image A1 corresponds to the reconstruction of Input~\#1, A8 to Input~\#2, H1 to Input~\#3, and H8 to Input~\#4. The rest of the images are produced by interpolating between the reconstructions in the latent space---for example, between A1 and A8. As can be seen in the figure, the latent space is well-behaved and even the glasses in Input~\#3 do not cause problems. We emphasize that compared to many GAN methods, the interpolations in \PIONEER can be done elegantly and without separate optimization stage needed for each input sample.

\subsection{LSUN Bedrooms}
\label{sec:LSUN-bedrooms}
The LSUN dataset \cite{Yu:2015} contains images of various categories in $256{\times}256$ resolution or higher. We choose the category of bedrooms, often used for testing generative models. For humans, comparing randomly generated samples is more difficult on this dataset than with faces, so quantitative metrics are important to separate between the subtle differences in quality and diversity of captured features.

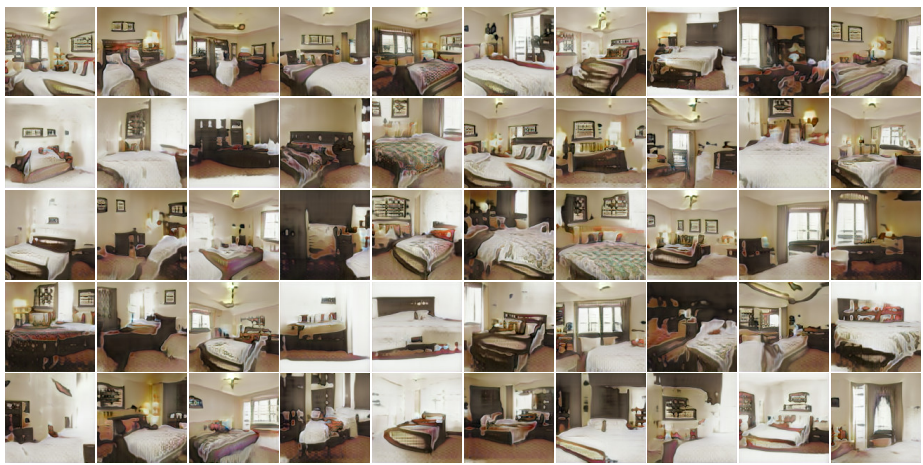
\begin{figure}[!t]
\centering\scriptsize
\setlength{\figurewidth}{.10\textwidth}
\setlength{\figureheight}{\figurewidth}
\newcommand{\figg}[1]{\includegraphics[width=.97\figurewidth]{./fig/fig5/final/000001_#1.jpg}}
\begin{tikzpicture}
  \foreach \i in {1,...,50}
    \node[draw=white,minimum size=\figurewidth,inner sep=0]
      (\i) at ({\figurewidth*mod(\i-1,10)},{-\figureheight*int((\i-1)/10)}) {\figg{\i}};
\end{tikzpicture}
\caption{Generated images of LSUN bedrooms at $128{\times}128$ resolution. Results can be compared to the image arrays in \cite{mao2016}, \cite{gulrajani2017}, and \cite{karras2017}.}
\label{fig:lsungen}
\vspace*{-1em}
\end{figure}

We ran the LSUN training similarly to \CelebA, but with only a single target resolution of $128{\times}128$. We present randomly generated samples from LSUN bedrooms (Fig.~\ref{fig:lsungen}) at $128{\times}128$ resolution. Comparing to the non-progressive GANs of \cite{gulrajani2017} and \cite{mao2016}, we see that \PIONEER output quality visually matches them, while falling slightly behind the fully generative ProgGAN, as expected. The FID of 37.50 was reached with no hyper-parameter tuning specific to LSUN.

For networks that support both inference and generation, we would not expect to achieve the same quality metrics as with purely generative models, so these results are not directly comparable.

\subsection{\Cifar}
\label{sec:cifar}
The \Cifar dataset contains 60,000 labeled images at $32{\times}32$ resolution, spanning 10 classes. As our method is fully supervised, we do not utilize the label information. During the training, we found that progressive growing seemed to provide no benefits. Therefore, we trained the \PIONEER model otherwise as normal, but started at $32{\times}32$ resolution and did not use progressive growing.

\begin{figure}[!t]
  \centering\scriptsize
  \setlength{\figurewidth}{.0625\textwidth}
  \setlength{\figureheight}{\figurewidth}
  \newcommand{\figg}[1]{\includegraphics[width=.97\figurewidth]{./fig/fig7/018750_511#1.jpg}}
  \begin{tikzpicture}
    \foreach \i in {1,...,96}
      \node[draw=white,minimum size=\figurewidth,inner sep=0]
        (\i) at ({\figurewidth*mod(\i-1,16)},{-\figureheight*int((\i-1)/16)}) {\figg{\i}};
  \end{tikzpicture}
  \caption{Generated images of \Cifar at $32{\times}32$ resolution.}
  \label{fig:cifar}
  \vspace*{-1em}
\end{figure}
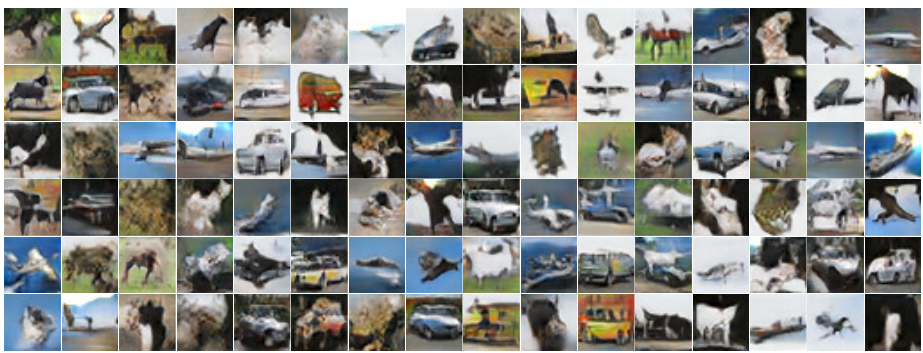

We used the same architecture, losses and algorithm as for the other datasets, instead of trying to optimize our approach to get the best results in \Cifar. We confirmed that the approach works, but it is not particularly suitable for this kind of a dataset without further modifications. Generated samples are provided in Figure~\ref{fig:cifar}. We believe that with some natural modifications, the model will be able to compete with GAN-based methods, but we leave this for our future work.

\section{Discussion and Conclusion}
\label{sec:discussion}
In this paper, we proposed a generative image autoencoder model that is trained with the combination of adversarial and reconstruction losses in sample and latent space, using progressive growing of generator and encoder networks, but without a separate GAN-like discriminator. We showed that this model can both generate sharp images---at least up to $256{\times}256$ resolution---and carry out inference on input images at least up to $128{\times}128$ resolution with sharp output, while having a simpler architecture than the state-of-the-art of purely generative models \cite{karras2017}. We demonstrated the inference via sample reconstruction and smooth interpolation in the latent space, and showed the overall generative capability by generating new random samples from the latent space and measuring the quality and diversity of the generated distribution against baselines.

We emphasize that evaluation of generative models is heavily dependent on the resolution, and there is a multitude of models that have been shown to work on $64{\times}64$ resolution, but not on $128{\times}128$ or above. Reaching higher resolutions is not only a matter of raw compute, but the model needs to be able to cope with the increasing information and be regularised suitably in order not to loose the representative power or become instable.

We found that training is more stable using spectral normalization, which also suits our non-GAN loss architecture and loss. The model provides image reconstruction results with larger image resolutions than previous state-of-the-art. Importantly, our model has only few hyper-parameters and is robust to train. The only hyper-parameter that typically needs to be tuned between datasets is the number of epochs spent on intermediate resolutions. Our results indicate that the GAN paradigm of a separate discriminator network may not be necessary for learning to infer and generate image data sets. GANs do currently remain the best option if one is only interested in generating random samples. Like GANs, our model is heavily based on the general idea of `adversarial' training, construed as setting the generator--encoder pair up with opposite gradients to each other with respect to the source of the data (that is, simulated vs.\ observed).

As Karras~\etal \cite{karras2017} point out for GANs, the principle of growing the network progressively may be more important than the specific loss function formulation. Likewise, even though the AGE formulation for the latent space loss metrics is relatively simple, we believe that there are many ways in which the encoder can be set up to achieve and exceed the results we have demonstrated here. 

In future work, we will also continue training the network to carry out faithful reconstructions at $256{\times}256$, $512{\times}512$, and $1024{\times}1024$ resolutions, omitted from this paper primarily due to the extensive amount of computation (or preferably, further optimization) required. We will also further investigate whether the \CelebAHQ dataset is sufficiently diverse for this purpose.

\subsubsection*{Acknowledgments}

We thank Tero Karras, Dmitry Ulyanov and Jaakko Lehtinen for fruitful discussions. We acknowledge the computational resources provided by the Aalto Science-IT project. Authors acknowledge funding from the Academy of Finland (grant numbers 308640 and 277685) and GenMind Ltd.

\clearpage

\phantomsection%
\addcontentsline{toc}{section}{References}
\begingroup
\bibliographystyle{splncs}
\bibliography{bibliography}

\begin{thebibliography}{10}

\bibitem{goodfellow2014}
Goodfellow, I.J., Pouget-Abadie, J., Mirza, M., Xu, B., Warde-Farley, D.,
  Ozair, S., Courville, A., Bengio, Y.:
\newblock Generative adversarial networks.
\newblock In: Advances in Neural Information Processing Systems (NIPS). (2014)

\bibitem{karras2017}
Karras, T., Aila, T., Laine, S., Lehtinen, J.:
\newblock Progressive growing of {GAN}s for improved quality, stability, and
  variation.
\newblock In: International Conference on Learning Representations (ICLR).
  (2018)

\bibitem{brock2016}
Brock, A., Lim, T., Ritchie, J.M., Weston, N.:
\newblock Neural photo editing with introspective adversarial networks.
\newblock In: International Conference on Learning Representations (ICLR).
  (2017)

\bibitem{kingma2014}
Kingma, D., Welling, M.:
\newblock Auto-encoding variational {B}ayes.
\newblock In: International Conference on Learning Representations (ICLR).
  (2014)

\bibitem{rezende2014}
Jimenez~Rezende, D., Mohamed, S., Wierstra, D.:
\newblock Stochastic backpropagation and approximate inference in deep
  generative models.
\newblock In: International Conference on Machine Learning (ICML). (2014)

\bibitem{larsen2015}
Boesen Lindbo~Larsen, A., Kaae~S{\o}nderby, S., Larochelle, H., Winther, O.:
\newblock Autoencoding beyond pixels using a learned similarity metric.
\newblock In: International Conference on Machine Learning (ICML). (2016)

\bibitem{makhzani2015}
Makhzani, A., Shlens, J., Jaitly, N., Goodfellow, I., Frey, B.:
\newblock Adversarial autoencoders.
\newblock arXiv preprint arXiv:1511.05644 (2015)

\bibitem{Donahue2017}
Donahue, J., {Kr\"ahenb\"uhl}, P., Darrell, T.:
\newblock Adversarial feature learning.
\newblock In: International Conference on Learning Representations (ICLR).
  (2017)

\bibitem{dumoulin2016D}
Dumoulin, V., Belghazi, I., Poole, B., Mastropietro, O., Lamb, A., Arjovsky,
  M., Courville, A.:
\newblock Adversarially learned inference.
\newblock In: International Conference on Learning Representations (ICLR).
  (2017)

\bibitem{ulyanov2017}
Ulyanov, D., Vedaldi, A., Lempitsky, V.:
\newblock It takes (only) two: Adversarial generator-encoder networks.
\newblock In: AAAI Conference on Artificial Intelligence. (2018)

\bibitem{Rosca2017}
Rosca, M., Lakshminarayanan, B., Warde-Farley, D., Mohamed, S.:
\newblock Variational approaches for auto-encoding generative adversarial
  networks.
\newblock arXiv preprint arXiv:1706.04987 (2017)

\bibitem{Zhou+Bovik+Sheikh+Simoncelli:2004}
Zhou, W., Bovik, A.C., Sheikh, H.R., Simoncelli, E.P.:
\newblock Image qualifty assessment: {F}rom error visibility to structural
  similarity.
\newblock IEEE Transactions on Image Processing \textbf{13} (2008)  600--612

\bibitem{miyato2018}
Miyato, T., Kataoka, T., Koyama, M., Yoshida, Y.:
\newblock Spectral normalization for generative adversarial networks.
\newblock In: International Conference on Learning Representations (ICLR).
  (2018)

\bibitem{gulrajani2017}
Gulrajani, I., Ahmed, F., Arjovsky, M., Dumoulin, V., Courville, A.C.:
\newblock Improved training of wasserstein {GAN}s.
\newblock In: Advances in Neural Information Processing Systems (NIPS).
\newblock (2017)

\bibitem{Liu:2015}
Liu, Z., Luo, P., Wang, X., Tang, X.:
\newblock Deep learning face attributes in the wild.
\newblock In: International Conference on Computer Vision (ICCV). (2015)

\bibitem{li2015}
Li, Y., Swersky, K., Zemel, R.:
\newblock Generative moment matching networks.
\newblock In: International Conference on Machine Learning (ICML). (2015)

\bibitem{Arjovsky2017b}
Arjovsky, M., Chintala, S., Bottou, L.:
\newblock Wasserstein generative adversarial networks.
\newblock In: International Conference on Machine Learning (ICML). (2017)

\bibitem{zhang2016}
Zhang, H., Xu, T., Li, H., Zhang, S., Wang, X., Huang, X., Metaxas, D.:
\newblock {StackGAN}: {T}ext to photo-realistic image synthesis with stacked
  generative adversarial networks.
\newblock In: International Conference on Computer Vision (ICCV). (2017)

\bibitem{RothLNH17}
Roth, K., Lucchi, A., Nowozin, S., Hofmann, T.:
\newblock Stabilizing training of generative adversarial networks through
  regularization.
\newblock In: Advances in Neural Information Processing Systems (NIPS). (2017)

\bibitem{qi2017}
Qi, G.J.:
\newblock Loss-sensitive generative adversarial networks on {L}ipschitz
  densities.
\newblock arXiv preprint arXiv:1701.06264 (2017)

\bibitem{mescheder2017}
Mescheder, L., Nowozin, S., Geiger, A.:
\newblock Adversarial variational {B}ayes: {U}nifying variational autoencoders
  and generative adversarial networks.
\newblock In: International Conference on Machine Learning (ICML). (2017)

\bibitem{tabor2018}
Tabor, J., Knop, S., Spurek, P., Podolak, I., Mazur, M., Jastrz\k{e}bski, S.:
\newblock Cramer--{W}old {AutoEncoder}.
\newblock arXiv preprint arXiv:1805.09235 (2018)

\bibitem{shrivastava2016}
Shrivastava, A., Pfister, T., Tuzel, O., Susskind, J., Wang, W., Webb, R.:
\newblock Learning from simulated and unsupervised images through adversarial
  training.
\newblock In: CVPR. (2017)

\bibitem{adam}
Kingma, D.P., Ba, J.:
\newblock Adam: {A} method for stochastic optimization.
\newblock In: International Conference on Learning Representations (ICLR).
  (2015)

\bibitem{Yu:2015}
Yu, F., Seff, A., Zhang, Y., Song, S., Funkhouser, T., Xiao, J.:
\newblock {LSUN}: {C}onstruction of a large-scale image dataset using deep
  learning with humans in the loop.
\newblock arXiv preprint arXiv:1506.03365 (2015)

\bibitem{ulyanovGithub}
Ulyanov, D., Vedaldi, A., Lempitsky, V.:
\newblock Adversarial generator-encoder networks.
\newblock \url{https://github.com/DmitryUlyanov/AGE} (2018) GitHub repository.

\bibitem{heusel2017}
Heusel, M., Ramsauer, H., Unterthiner, T., Nessler, B., Hochreiter, S.:
\newblock {GAN}s trained by a two time-scale update rule converge to a local
  {N}ash equilibrium.
\newblock In: Advances in Neural Information Processing Systems (NIPS). (2017)

\bibitem{rabin2015}
Rabin, J., Peyr\'e, G., Bernot, M.:
\newblock Wasserstein barycenter and its application to texture mixing.
\newblock In: International Conference on Scale Space and Variational Methods
  in Computer Vision. (2011)

\bibitem{rosca2018}
Rosca, M., Lakshminarayanan, B., Mohamed, S.:
\newblock Distribution matching in variational inference.
\newblock arXiv preprint arXiv:1802.06847 (2018)

\bibitem{radford2015}
Radford, A., Metz, L., Chintala, S.:
\newblock Unsupervised representation learning with deep convolutional
  generative adversarial networks.
\newblock In: International Conference on Learning Representations (ICLR).
  (2016)

\bibitem{lipton2017}
Lipton, Z.C., Tripathi, S.:
\newblock Precise recovery of latent vectors from generative adversarial
  networks.
\newblock In: International Conference on Learning Representations (ICLR).
  (2017)

\bibitem{luo2017}
Luo, J., Xu, Y., Tang, C., Lv, J.:
\newblock Learning inverse mapping by autoencoder based generative adversarial
  nets.
\newblock In: Neural Information Processing (ICONIP). Volume 10635 of Lecture
  Notes in Computer Science. (2017)

\bibitem{creswell2016}
Creswell, A., Bharath, A.A.:
\newblock Inverting the generator of a generative adversarial network.
\newblock In: NIPS 2016 Workshop on Adversarial Training. (2016)

\bibitem{mao2016}
Mao, X., Li, Q., Xie, H., Lau, R.Y.K., Wang, Z., Smolley, S.P.:
\newblock Least squares generative adversarial networks.
\newblock In: International Conference on Computer Vision (ICCV). (2017)

\end{thebibliography}
\endgroup

\clearpage\appendix
\setcounter{section}{0}
\appendixtitle{Supplementary Material for Pioneer Networks: \mbox{Progressively Growing  Generative Autoencoder}}
\pagestyle{empty}

\vspace*{3em}

\noindent
In this supplementary, we provide additional experiment figures that provide a broader overview on how the proposed method performs on \CelebA and \CelebAHQ. We also include generated samples from the ALI and AGE methods.

\begin{figure}[!h]
\centering\scriptsize
\setlength{\figurewidth}{.09\textwidth}
\setlength{\figureheight}{\figurewidth}
\newcommand\x{x}
\begin{tikzpicture}
  \foreach \i in {0,...,7}
    \foreach \j in {0,...,7}
      \node[draw=white,fill=black!20,minimum size=\figurewidth,inner sep=0]
        (\i) at ({\figurewidth*\i},{-\figureheight*\j})
        {\includegraphics[width=\figurewidth]{./supplement/pine_interpolations/interpolations_5_480002_1.0_\j\x\i.jpg}};

  \foreach \i in {0,...,7}
     \node at ({\figurewidth*\i},{0.7\figureheight}) {\bf\pgfmathparse{int(\i+1)}\pgfmathresult};

  \def\myarray{{"A","B","C","D","E","F","G","H","I"}}
  \foreach \i in {0,...,7}
     \node at ({-0.7\figureheight},{-\figurewidth*\i}) {\bf\pgfmathparse{\myarray[\i]}\pgfmathresult};

  \node[inner sep=0] at 
    ({-1.5*\figurewidth},{-0\figureheight}) 
    {\includegraphics[width=1\figurewidth]{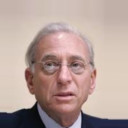}};
  \node[inner sep=0] at 
    ({-1.5*\figurewidth+10*\figurewidth},{-0\figureheight}) 
    {\includegraphics[width=1\figurewidth]{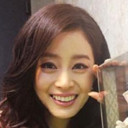}};        
  \node[inner sep=0] at 
    ({-1.5*\figurewidth},{-0\figureheight-7*\figureheight}) 
    {\includegraphics[width=1\figurewidth]{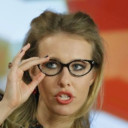}};
  \node[inner sep=0] at 
    ({-1.5*\figurewidth+10*\figurewidth},{-0\figureheight-7*\figureheight}) 
    {\includegraphics[width=1\figurewidth]{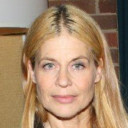}};

  \node[inner sep=0] at 
    ({-1.5*\figurewidth},{-0.75\figureheight}) {\bf In \#1};
  \node[inner sep=0] at 
    ({-1.5*\figurewidth+10*\figureheight},{-0.75\figureheight}) {\bf In \#2};
  \node[inner sep=0] at 
    ({-1.5*\figurewidth},{+0.75\figureheight-7*\figureheight}) {\bf In \#3};
  \node[inner sep=0] at 
    ({-1.5*\figurewidth+10*\figureheight},{+0.75\figureheight-7*\figureheight}) {\bf In \#4};
    
\end{tikzpicture}
\caption{\PIONEER interpolation example on test set input images at $128{\times}128$ resolution.} 
\end{figure}

\begin{figure}[!t]
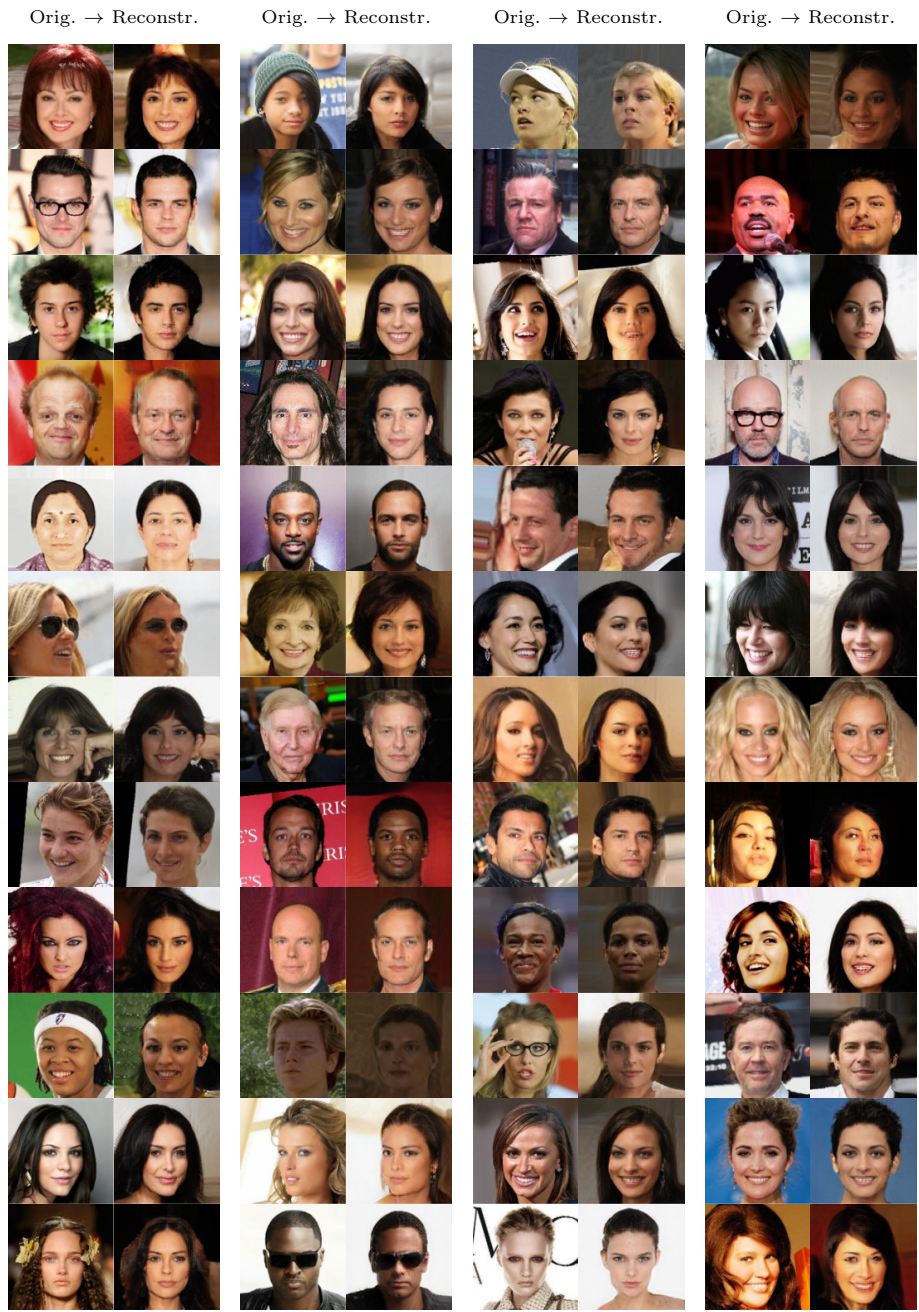

  \centering\scriptsize
  \setlength{\figurewidth}{0.115\textwidth}
  \setlength{\figureheight}{\figurewidth}
  \begin{tikzpicture}
    \foreach \i in {0,...,47} {%
      \node[draw=white,fill=black!20,minimum size=\figurewidth,inner sep=0pt]
        (\i) at ({2.2*\figurewidth*int((\i)/12)},{-\figureheight*mod(\i,12)})
        {\includegraphics[width=\figurewidth]{./supplement/pine_reconstruct_128x128/\i_orig.jpg}};
      \node[draw=white,fill=black!20,minimum size=\figurewidth,inner sep=0pt]
        (\i) at ({2.2*\figurewidth*int((\i)/12)+\figurewidth},{-\figureheight*mod(\i,12)})
        {\includegraphics[width=\figurewidth]{./supplement/pine_reconstruct_128x128/\i_pine.jpg}};        
      }
      
    \foreach \i in {0,...,3} 
      \node at ({2.2*\figurewidth*\i+0.5*\figurewidth},.75\figureheight) {Orig.\ $\to$ Reconstr.};

  \end{tikzpicture}
  \caption{Examples of \PIONEER network reconstruction (\CelebA) quality in $128{\times}128$ resolution.}
\end{figure}

\begin{figure}[!h]
  \centering\scriptsize
  \setlength{\figurewidth}{.18\textwidth}
  \setlength{\figureheight}{\figurewidth}
  \begin{tikzpicture}

    \tikzstyle{fig} = [draw=white,minimum size=\figurewidth,inner sep=0pt]

    \newcommand{\figg}[1]{\includegraphics[width=.97\figurewidth]{./supplement/pine_random_256x256/#1.jpg}};

    \node[fig] at ({\figurewidth*0},{-\figureheight*0}) {\figg{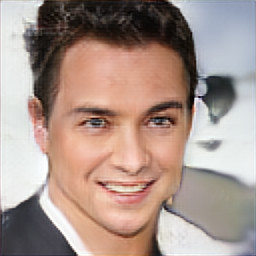}};
    \node[fig] at ({\figurewidth*1},{-\figureheight*0}) {\figg{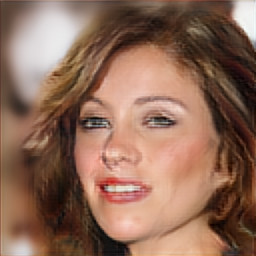}};
    \node[fig] at ({\figurewidth*2},{-\figureheight*0}) {\figg{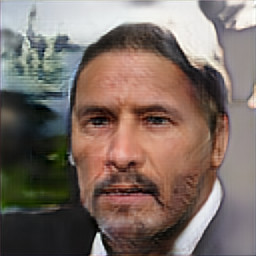}};
    \node[fig] at ({\figurewidth*3},{-\figureheight*0}) {\figg{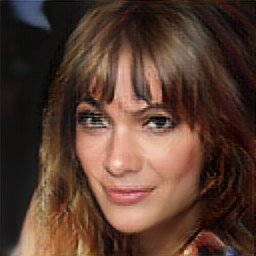}};
    \node[fig] at ({\figurewidth*4},{-\figureheight*0}) {\figg{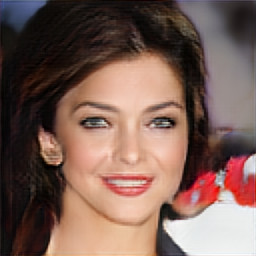}};
    \node[fig] at ({\figurewidth*0},{-\figureheight*1}) {\figg{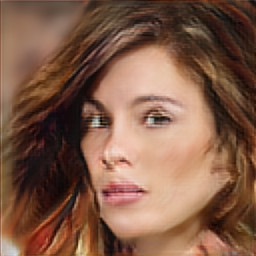}};
    \node[fig] at ({\figurewidth*1},{-\figureheight*1}) {\figg{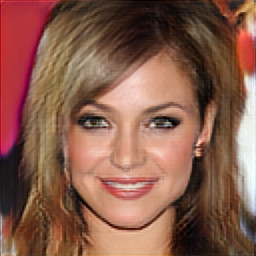}};
    \node[fig] at ({\figurewidth*2},{-\figureheight*1}) {\figg{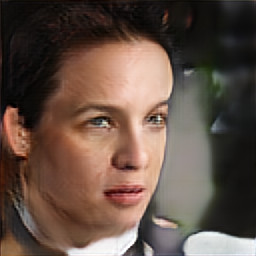}};
    \node[fig] at ({\figurewidth*3},{-\figureheight*1}) {\figg{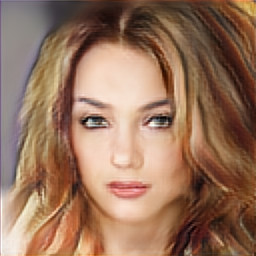}};
    \node[fig] at ({\figurewidth*4},{-\figureheight*1}) {\figg{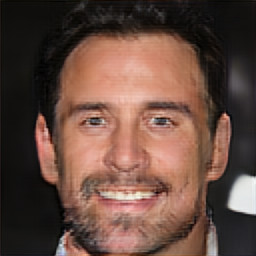}};
    \node[fig] at ({\figurewidth*0},{-\figureheight*2}) {\figg{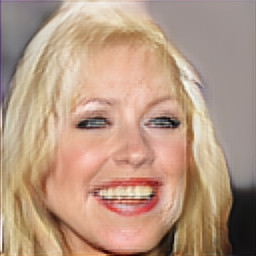}};
    \node[fig] at ({\figurewidth*1},{-\figureheight*2}) {\figg{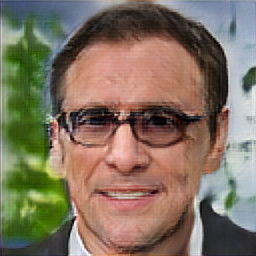}};
    \node[fig] at ({\figurewidth*2},{-\figureheight*2}) {\figg{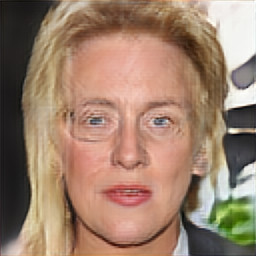}};
    \node[fig] at ({\figurewidth*3},{-\figureheight*2}) {\figg{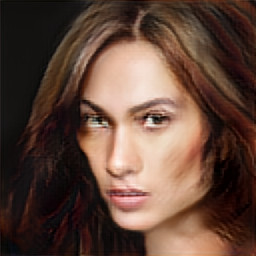}};
    \node[fig] at ({\figurewidth*4},{-\figureheight*2}) {\figg{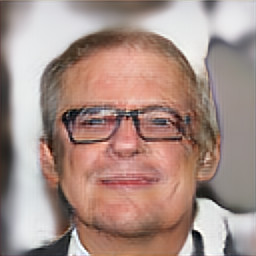}};
    \node[fig] at ({\figurewidth*0},{-\figureheight*3}) {\figg{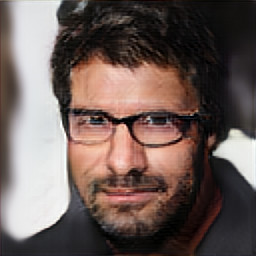}};
    \node[fig] at ({\figurewidth*1},{-\figureheight*3}) {\figg{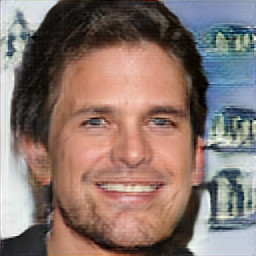}};
    \node[fig] at ({\figurewidth*2},{-\figureheight*3}) {\figg{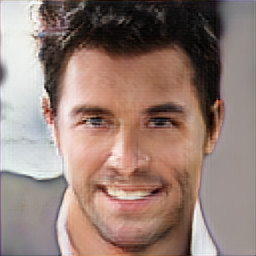}};
    \node[fig] at ({\figurewidth*3},{-\figureheight*3}) {\figg{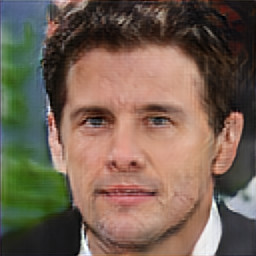}};
    \node[fig] at ({\figurewidth*4},{-\figureheight*3}) {\figg{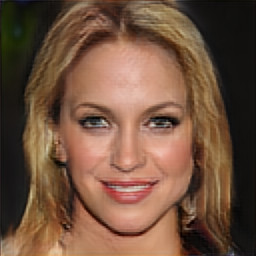}};
    \node[fig] at ({\figurewidth*0},{-\figureheight*4}) {\figg{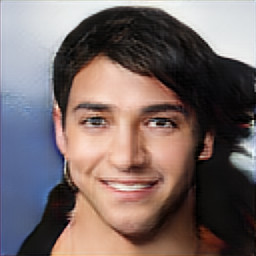}};
    \node[fig] at ({\figurewidth*1},{-\figureheight*4}) {\figg{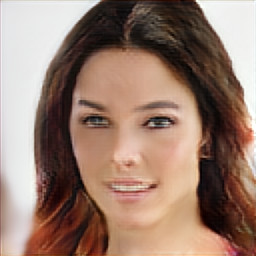}};
    \node[fig] at ({\figurewidth*2},{-\figureheight*4}) {\figg{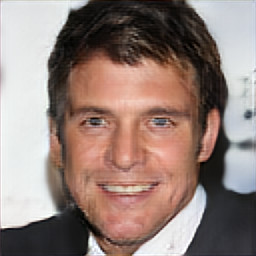}};
    \node[fig] at ({\figurewidth*3},{-\figureheight*4}) {\figg{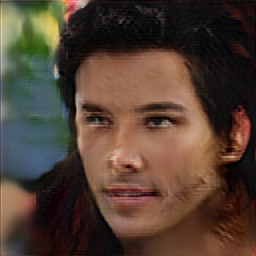}};
    \node[fig] at ({\figurewidth*4},{-\figureheight*4}) {\figg{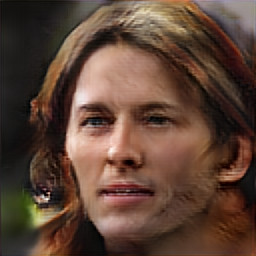}};
    \node[fig] at ({\figurewidth*0},{-\figureheight*5}) {\figg{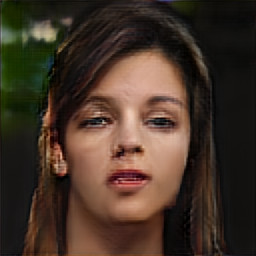}};
    \node[fig] at ({\figurewidth*1},{-\figureheight*5}) {\figg{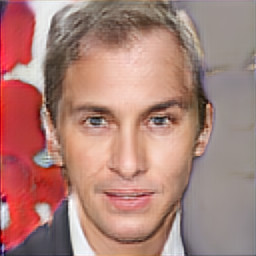}};
    \node[fig] at ({\figurewidth*2},{-\figureheight*5}) {\figg{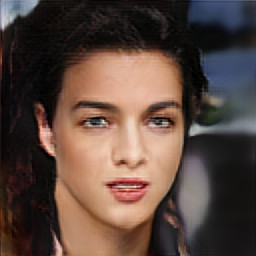}};
    \node[fig] at ({\figurewidth*3},{-\figureheight*5}) {\figg{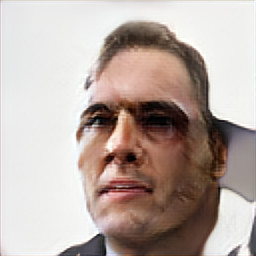}};
    \node[fig] at ({\figurewidth*4},{-\figureheight*5}) {\figg{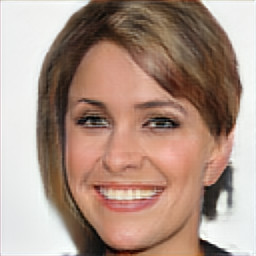}};
    \node[fig] at ({\figurewidth*0},{-\figureheight*6}) {\figg{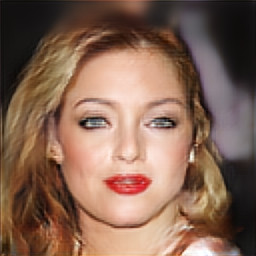}};
    \node[fig] at ({\figurewidth*1},{-\figureheight*6}) {\figg{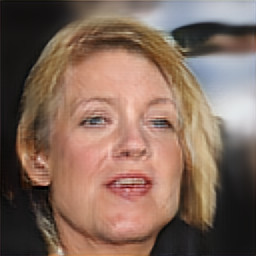}};
    \node[fig] at ({\figurewidth*2},{-\figureheight*6}) {\figg{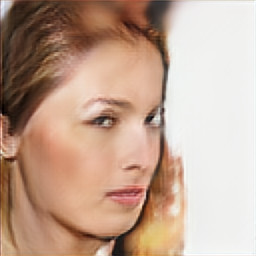}};
    \node[fig] at ({\figurewidth*3},{-\figureheight*6}) {\figg{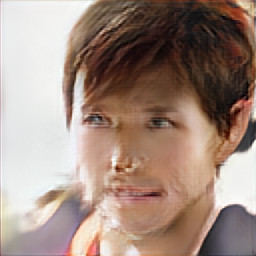}};
    \node[fig] at ({\figurewidth*4},{-\figureheight*6}) {\figg{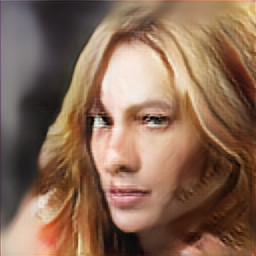}};
    \node[fig] at ({\figurewidth*0},{-\figureheight*7}) {\figg{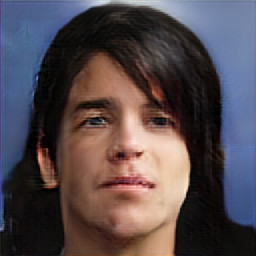}};
    \node[fig] at ({\figurewidth*1},{-\figureheight*7}) {\figg{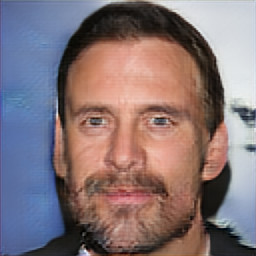}};
    \node[fig] at ({\figurewidth*2},{-\figureheight*7}) {\figg{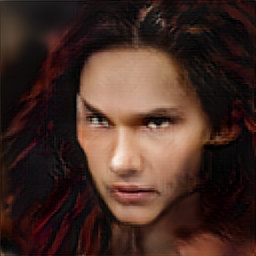}};
    \node[fig] at ({\figurewidth*3},{-\figureheight*7}) {\figg{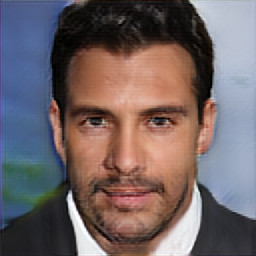}};
    \node[fig] at ({\figurewidth*4},{-\figureheight*7}) {\figg{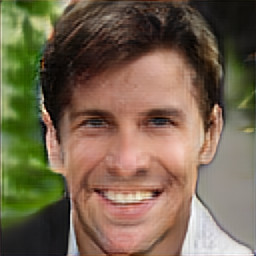}};

  \end{tikzpicture}
  \caption{\PIONEER random samples (\CelebAHQ) at $256{\times}256$ resolution.}
\end{figure}

\begin{figure}[!h]
  \centering\scriptsize
  \setlength{\figurewidth}{.09\textwidth}
  \setlength{\figureheight}{\figurewidth}
  \begin{tikzpicture}

    \tikzstyle{fig} = [draw=white,minimum size=\figurewidth,inner sep=0pt]

    \newcommand{\figg}[1]{\includegraphics[width=.97\figurewidth]{./supplement/pine_random_128x128/#1.jpg}};

    \node[fig] at ({\figurewidth*0},{-\figureheight*0}) {\figg{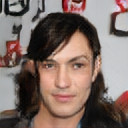}};
    \node[fig] at ({\figurewidth*1},{-\figureheight*0}) {\figg{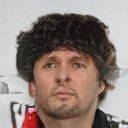}};
    \node[fig] at ({\figurewidth*2},{-\figureheight*0}) {\figg{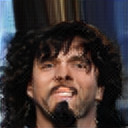}};
    \node[fig] at ({\figurewidth*3},{-\figureheight*0}) {\figg{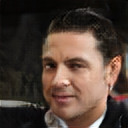}};
    \node[fig] at ({\figurewidth*4},{-\figureheight*0}) {\figg{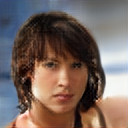}};
    \node[fig] at ({\figurewidth*5},{-\figureheight*0}) {\figg{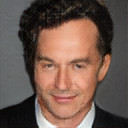}};
    \node[fig] at ({\figurewidth*6},{-\figureheight*0}) {\figg{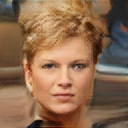}};
    \node[fig] at ({\figurewidth*7},{-\figureheight*0}) {\figg{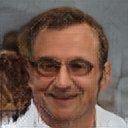}};
    \node[fig] at ({\figurewidth*8},{-\figureheight*0}) {\figg{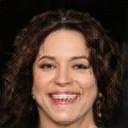}};
    \node[fig] at ({\figurewidth*9},{-\figureheight*0}) {\figg{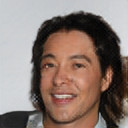}};
    \node[fig] at ({\figurewidth*0},{-\figureheight*1}) {\figg{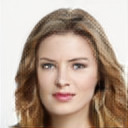}};
    \node[fig] at ({\figurewidth*1},{-\figureheight*1}) {\figg{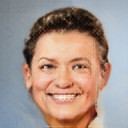}};
    \node[fig] at ({\figurewidth*2},{-\figureheight*1}) {\figg{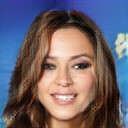}};
    \node[fig] at ({\figurewidth*3},{-\figureheight*1}) {\figg{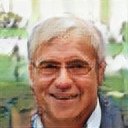}};
    \node[fig] at ({\figurewidth*4},{-\figureheight*1}) {\figg{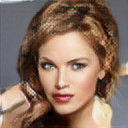}};
    \node[fig] at ({\figurewidth*5},{-\figureheight*1}) {\figg{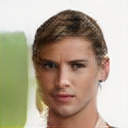}};
    \node[fig] at ({\figurewidth*6},{-\figureheight*1}) {\figg{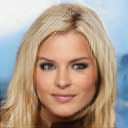}};
    \node[fig] at ({\figurewidth*7},{-\figureheight*1}) {\figg{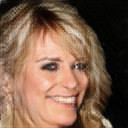}};
    \node[fig] at ({\figurewidth*8},{-\figureheight*1}) {\figg{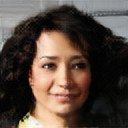}};
    \node[fig] at ({\figurewidth*9},{-\figureheight*1}) {\figg{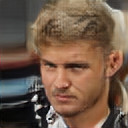}};
    \node[fig] at ({\figurewidth*0},{-\figureheight*2}) {\figg{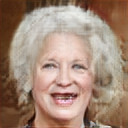}};
    \node[fig] at ({\figurewidth*1},{-\figureheight*2}) {\figg{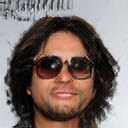}};
    \node[fig] at ({\figurewidth*2},{-\figureheight*2}) {\figg{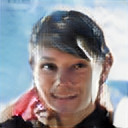}};
    \node[fig] at ({\figurewidth*3},{-\figureheight*2}) {\figg{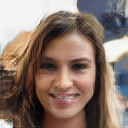}};
    \node[fig] at ({\figurewidth*4},{-\figureheight*2}) {\figg{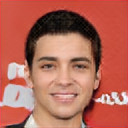}};
    \node[fig] at ({\figurewidth*5},{-\figureheight*2}) {\figg{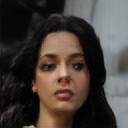}};
    \node[fig] at ({\figurewidth*6},{-\figureheight*2}) {\figg{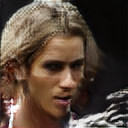}};
    \node[fig] at ({\figurewidth*7},{-\figureheight*2}) {\figg{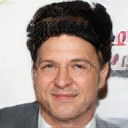}};
    \node[fig] at ({\figurewidth*8},{-\figureheight*2}) {\figg{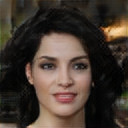}};
    \node[fig] at ({\figurewidth*9},{-\figureheight*2}) {\figg{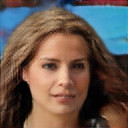}};
    \node[fig] at ({\figurewidth*0},{-\figureheight*3}) {\figg{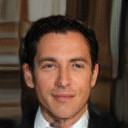}};
    \node[fig] at ({\figurewidth*1},{-\figureheight*3}) {\figg{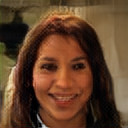}};
    \node[fig] at ({\figurewidth*2},{-\figureheight*3}) {\figg{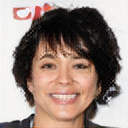}};
    \node[fig] at ({\figurewidth*3},{-\figureheight*3}) {\figg{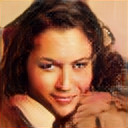}};
    \node[fig] at ({\figurewidth*4},{-\figureheight*3}) {\figg{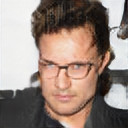}};
    \node[fig] at ({\figurewidth*5},{-\figureheight*3}) {\figg{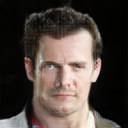}};
    \node[fig] at ({\figurewidth*6},{-\figureheight*3}) {\figg{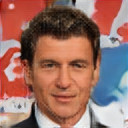}};
    \node[fig] at ({\figurewidth*7},{-\figureheight*3}) {\figg{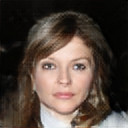}};
    \node[fig] at ({\figurewidth*8},{-\figureheight*3}) {\figg{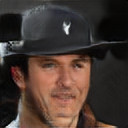}};
    \node[fig] at ({\figurewidth*9},{-\figureheight*3}) {\figg{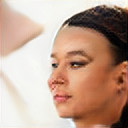}};
    \node[fig] at ({\figurewidth*0},{-\figureheight*4}) {\figg{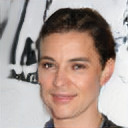}};
    \node[fig] at ({\figurewidth*1},{-\figureheight*4}) {\figg{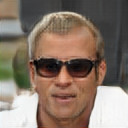}};
    \node[fig] at ({\figurewidth*2},{-\figureheight*4}) {\figg{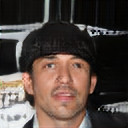}};
    \node[fig] at ({\figurewidth*3},{-\figureheight*4}) {\figg{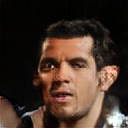}};
    \node[fig] at ({\figurewidth*4},{-\figureheight*4}) {\figg{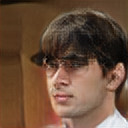}};
    \node[fig] at ({\figurewidth*5},{-\figureheight*4}) {\figg{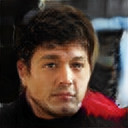}};
    \node[fig] at ({\figurewidth*6},{-\figureheight*4}) {\figg{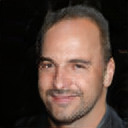}};
    \node[fig] at ({\figurewidth*7},{-\figureheight*4}) {\figg{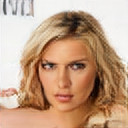}};
    \node[fig] at ({\figurewidth*8},{-\figureheight*4}) {\figg{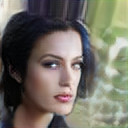}};
    \node[fig] at ({\figurewidth*9},{-\figureheight*4}) {\figg{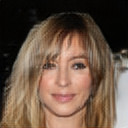}};
    \node[fig] at ({\figurewidth*0},{-\figureheight*5}) {\figg{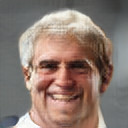}};
    \node[fig] at ({\figurewidth*1},{-\figureheight*5}) {\figg{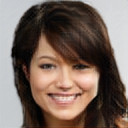}};
    \node[fig] at ({\figurewidth*2},{-\figureheight*5}) {\figg{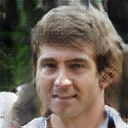}};
    \node[fig] at ({\figurewidth*3},{-\figureheight*5}) {\figg{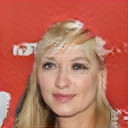}};
    \node[fig] at ({\figurewidth*4},{-\figureheight*5}) {\figg{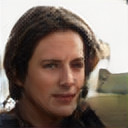}};
    \node[fig] at ({\figurewidth*5},{-\figureheight*5}) {\figg{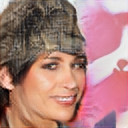}};
    \node[fig] at ({\figurewidth*6},{-\figureheight*5}) {\figg{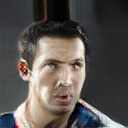}};
    \node[fig] at ({\figurewidth*7},{-\figureheight*5}) {\figg{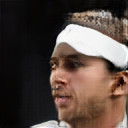}};
    \node[fig] at ({\figurewidth*8},{-\figureheight*5}) {\figg{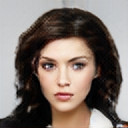}};
    \node[fig] at ({\figurewidth*9},{-\figureheight*5}) {\figg{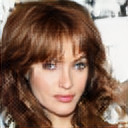}};

  \end{tikzpicture}
  \caption{\PIONEER random samples (\CelebA) at $128{\times}128$ resolution.}
\end{figure}

\begin{figure}[!h]
  \centering\scriptsize
  \setlength{\figurewidth}{.09\textwidth}
  \setlength{\figureheight}{\figurewidth}
  \begin{tikzpicture}

    \tikzstyle{fig} = [draw=white,minimum size=\figurewidth,inner sep=0pt]

    \newcommand{\figg}[1]{\includegraphics[width=.97\figurewidth]{./supplement/age_random_128x128/#1.jpg}};

    \node[fig] at ({\figurewidth*0},{-\figureheight*0}) {\figg{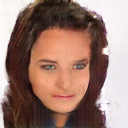}};
    \node[fig] at ({\figurewidth*1},{-\figureheight*0}) {\figg{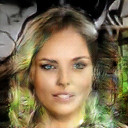}};
    \node[fig] at ({\figurewidth*2},{-\figureheight*0}) {\figg{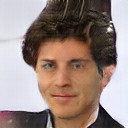}};
    \node[fig] at ({\figurewidth*3},{-\figureheight*0}) {\figg{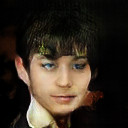}};
    \node[fig] at ({\figurewidth*4},{-\figureheight*0}) {\figg{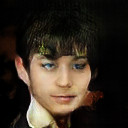}};
    \node[fig] at ({\figurewidth*5},{-\figureheight*0}) {\figg{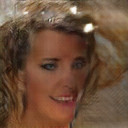}};
    \node[fig] at ({\figurewidth*6},{-\figureheight*0}) {\figg{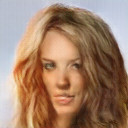}};
    \node[fig] at ({\figurewidth*7},{-\figureheight*0}) {\figg{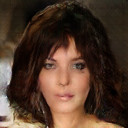}};
    \node[fig] at ({\figurewidth*8},{-\figureheight*0}) {\figg{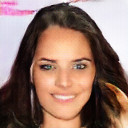}};
    \node[fig] at ({\figurewidth*9},{-\figureheight*0}) {\figg{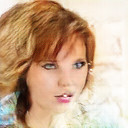}};
    \node[fig] at ({\figurewidth*0},{-\figureheight*1}) {\figg{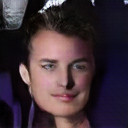}};
    \node[fig] at ({\figurewidth*1},{-\figureheight*1}) {\figg{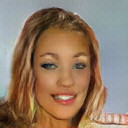}};
    \node[fig] at ({\figurewidth*2},{-\figureheight*1}) {\figg{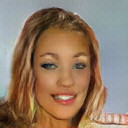}};
    \node[fig] at ({\figurewidth*3},{-\figureheight*1}) {\figg{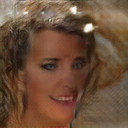}};
    \node[fig] at ({\figurewidth*4},{-\figureheight*1}) {\figg{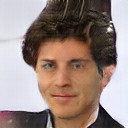}};
    \node[fig] at ({\figurewidth*5},{-\figureheight*1}) {\figg{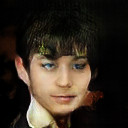}};
    \node[fig] at ({\figurewidth*6},{-\figureheight*1}) {\figg{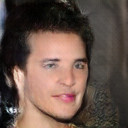}};
    \node[fig] at ({\figurewidth*7},{-\figureheight*1}) {\figg{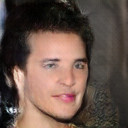}};
    \node[fig] at ({\figurewidth*8},{-\figureheight*1}) {\figg{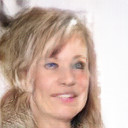}};
    \node[fig] at ({\figurewidth*9},{-\figureheight*1}) {\figg{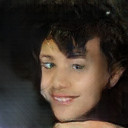}};
    \node[fig] at ({\figurewidth*0},{-\figureheight*2}) {\figg{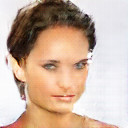}};
    \node[fig] at ({\figurewidth*1},{-\figureheight*2}) {\figg{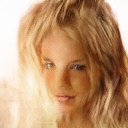}};
    \node[fig] at ({\figurewidth*2},{-\figureheight*2}) {\figg{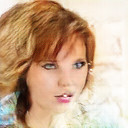}};
    \node[fig] at ({\figurewidth*3},{-\figureheight*2}) {\figg{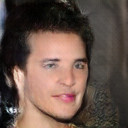}};
    \node[fig] at ({\figurewidth*4},{-\figureheight*2}) {\figg{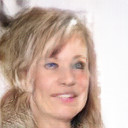}};
    \node[fig] at ({\figurewidth*5},{-\figureheight*2}) {\figg{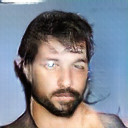}};
    \node[fig] at ({\figurewidth*6},{-\figureheight*2}) {\figg{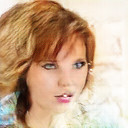}};
    \node[fig] at ({\figurewidth*7},{-\figureheight*2}) {\figg{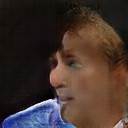}};
    \node[fig] at ({\figurewidth*8},{-\figureheight*2}) {\figg{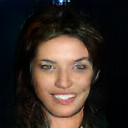}};
    \node[fig] at ({\figurewidth*9},{-\figureheight*2}) {\figg{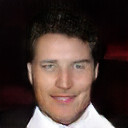}};
    \node[fig] at ({\figurewidth*0},{-\figureheight*3}) {\figg{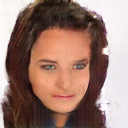}};
    \node[fig] at ({\figurewidth*1},{-\figureheight*3}) {\figg{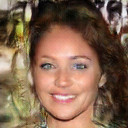}};
    \node[fig] at ({\figurewidth*2},{-\figureheight*3}) {\figg{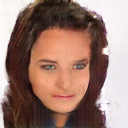}};
    \node[fig] at ({\figurewidth*3},{-\figureheight*3}) {\figg{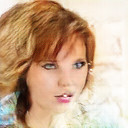}};
    \node[fig] at ({\figurewidth*4},{-\figureheight*3}) {\figg{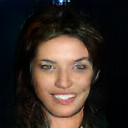}};
    \node[fig] at ({\figurewidth*5},{-\figureheight*3}) {\figg{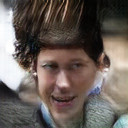}};
    \node[fig] at ({\figurewidth*6},{-\figureheight*3}) {\figg{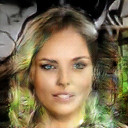}};
    \node[fig] at ({\figurewidth*7},{-\figureheight*3}) {\figg{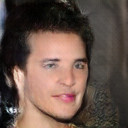}};
    \node[fig] at ({\figurewidth*8},{-\figureheight*3}) {\figg{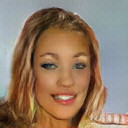}};
    \node[fig] at ({\figurewidth*9},{-\figureheight*3}) {\figg{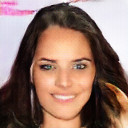}};
    \node[fig] at ({\figurewidth*0},{-\figureheight*4}) {\figg{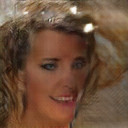}};
    \node[fig] at ({\figurewidth*1},{-\figureheight*4}) {\figg{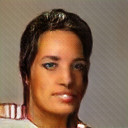}};
    \node[fig] at ({\figurewidth*2},{-\figureheight*4}) {\figg{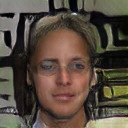}};
    \node[fig] at ({\figurewidth*3},{-\figureheight*4}) {\figg{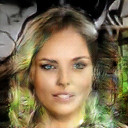}};
    \node[fig] at ({\figurewidth*4},{-\figureheight*4}) {\figg{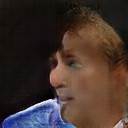}};
    \node[fig] at ({\figurewidth*5},{-\figureheight*4}) {\figg{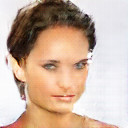}};
    \node[fig] at ({\figurewidth*6},{-\figureheight*4}) {\figg{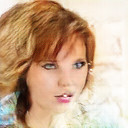}};
    \node[fig] at ({\figurewidth*7},{-\figureheight*4}) {\figg{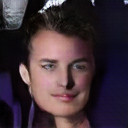}};
    \node[fig] at ({\figurewidth*8},{-\figureheight*4}) {\figg{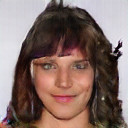}};
    \node[fig] at ({\figurewidth*9},{-\figureheight*4}) {\figg{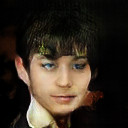}};
    \node[fig] at ({\figurewidth*0},{-\figureheight*5}) {\figg{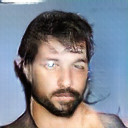}};
    \node[fig] at ({\figurewidth*1},{-\figureheight*5}) {\figg{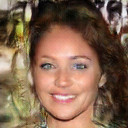}};
    \node[fig] at ({\figurewidth*2},{-\figureheight*5}) {\figg{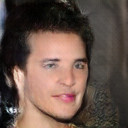}};
    \node[fig] at ({\figurewidth*3},{-\figureheight*5}) {\figg{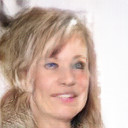}};
    \node[fig] at ({\figurewidth*4},{-\figureheight*5}) {\figg{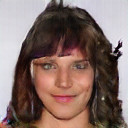}};
    \node[fig] at ({\figurewidth*5},{-\figureheight*5}) {\figg{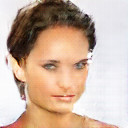}};
    \node[fig] at ({\figurewidth*6},{-\figureheight*5}) {\figg{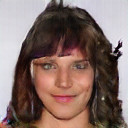}};
    \node[fig] at ({\figurewidth*7},{-\figureheight*5}) {\figg{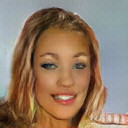}};
    \node[fig] at ({\figurewidth*8},{-\figureheight*5}) {\figg{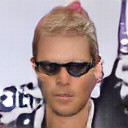}};
    \node[fig] at ({\figurewidth*9},{-\figureheight*5}) {\figg{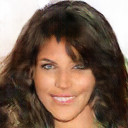}};

  \end{tikzpicture}
  \caption{AGE random samples (\CelebA) at $128{\times}128$ resolution.}
\end{figure}

\begin{figure}[!h]
  \centering\scriptsize
  \setlength{\figurewidth}{.09\textwidth}
  \setlength{\figureheight}{\figurewidth}
  \begin{tikzpicture}

    \tikzstyle{fig} = [draw=white,minimum size=\figurewidth,inner sep=0pt]

    \newcommand{\figg}[1]{\includegraphics[width=.97\figurewidth]{./supplement/age_random_64x64/#1.jpg}};

    \node[fig] at ({\figurewidth*0},{-\figureheight*0}) {\figg{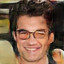}};
    \node[fig] at ({\figurewidth*1},{-\figureheight*0}) {\figg{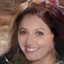}};
    \node[fig] at ({\figurewidth*2},{-\figureheight*0}) {\figg{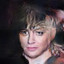}};
    \node[fig] at ({\figurewidth*3},{-\figureheight*0}) {\figg{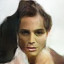}};
    \node[fig] at ({\figurewidth*4},{-\figureheight*0}) {\figg{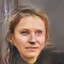}};
    \node[fig] at ({\figurewidth*5},{-\figureheight*0}) {\figg{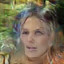}};
    \node[fig] at ({\figurewidth*6},{-\figureheight*0}) {\figg{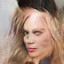}};
    \node[fig] at ({\figurewidth*7},{-\figureheight*0}) {\figg{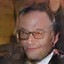}};
    \node[fig] at ({\figurewidth*8},{-\figureheight*0}) {\figg{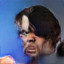}};
    \node[fig] at ({\figurewidth*9},{-\figureheight*0}) {\figg{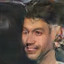}};
    \node[fig] at ({\figurewidth*0},{-\figureheight*1}) {\figg{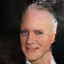}};
    \node[fig] at ({\figurewidth*1},{-\figureheight*1}) {\figg{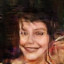}};
    \node[fig] at ({\figurewidth*2},{-\figureheight*1}) {\figg{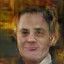}};
    \node[fig] at ({\figurewidth*3},{-\figureheight*1}) {\figg{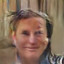}};
    \node[fig] at ({\figurewidth*4},{-\figureheight*1}) {\figg{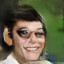}};
    \node[fig] at ({\figurewidth*5},{-\figureheight*1}) {\figg{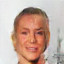}};
    \node[fig] at ({\figurewidth*6},{-\figureheight*1}) {\figg{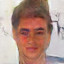}};
    \node[fig] at ({\figurewidth*7},{-\figureheight*1}) {\figg{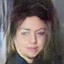}};
    \node[fig] at ({\figurewidth*8},{-\figureheight*1}) {\figg{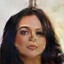}};
    \node[fig] at ({\figurewidth*9},{-\figureheight*1}) {\figg{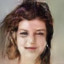}};
    \node[fig] at ({\figurewidth*0},{-\figureheight*2}) {\figg{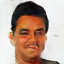}};
    \node[fig] at ({\figurewidth*1},{-\figureheight*2}) {\figg{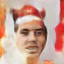}};
    \node[fig] at ({\figurewidth*2},{-\figureheight*2}) {\figg{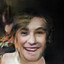}};
    \node[fig] at ({\figurewidth*3},{-\figureheight*2}) {\figg{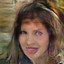}};
    \node[fig] at ({\figurewidth*4},{-\figureheight*2}) {\figg{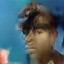}};
    \node[fig] at ({\figurewidth*5},{-\figureheight*2}) {\figg{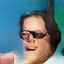}};
    \node[fig] at ({\figurewidth*6},{-\figureheight*2}) {\figg{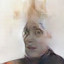}};
    \node[fig] at ({\figurewidth*7},{-\figureheight*2}) {\figg{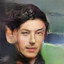}};
    \node[fig] at ({\figurewidth*8},{-\figureheight*2}) {\figg{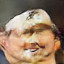}};
    \node[fig] at ({\figurewidth*9},{-\figureheight*2}) {\figg{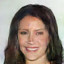}};
    \node[fig] at ({\figurewidth*0},{-\figureheight*3}) {\figg{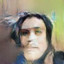}};
    \node[fig] at ({\figurewidth*1},{-\figureheight*3}) {\figg{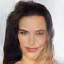}};
    \node[fig] at ({\figurewidth*2},{-\figureheight*3}) {\figg{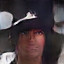}};
    \node[fig] at ({\figurewidth*3},{-\figureheight*3}) {\figg{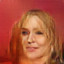}};
    \node[fig] at ({\figurewidth*4},{-\figureheight*3}) {\figg{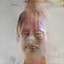}};
    \node[fig] at ({\figurewidth*5},{-\figureheight*3}) {\figg{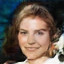}};
    \node[fig] at ({\figurewidth*6},{-\figureheight*3}) {\figg{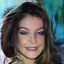}};
    \node[fig] at ({\figurewidth*7},{-\figureheight*3}) {\figg{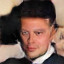}};
    \node[fig] at ({\figurewidth*8},{-\figureheight*3}) {\figg{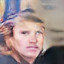}};
    \node[fig] at ({\figurewidth*9},{-\figureheight*3}) {\figg{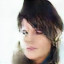}};
    \node[fig] at ({\figurewidth*0},{-\figureheight*4}) {\figg{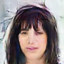}};
    \node[fig] at ({\figurewidth*1},{-\figureheight*4}) {\figg{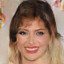}};
    \node[fig] at ({\figurewidth*2},{-\figureheight*4}) {\figg{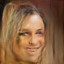}};
    \node[fig] at ({\figurewidth*3},{-\figureheight*4}) {\figg{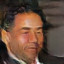}};
    \node[fig] at ({\figurewidth*4},{-\figureheight*4}) {\figg{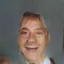}};
    \node[fig] at ({\figurewidth*5},{-\figureheight*4}) {\figg{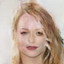}};
    \node[fig] at ({\figurewidth*6},{-\figureheight*4}) {\figg{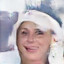}};
    \node[fig] at ({\figurewidth*7},{-\figureheight*4}) {\figg{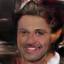}};
    \node[fig] at ({\figurewidth*8},{-\figureheight*4}) {\figg{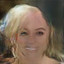}};
    \node[fig] at ({\figurewidth*9},{-\figureheight*4}) {\figg{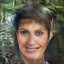}};
    \node[fig] at ({\figurewidth*0},{-\figureheight*5}) {\figg{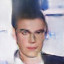}};
    \node[fig] at ({\figurewidth*1},{-\figureheight*5}) {\figg{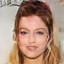}};
    \node[fig] at ({\figurewidth*2},{-\figureheight*5}) {\figg{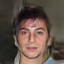}};
    \node[fig] at ({\figurewidth*3},{-\figureheight*5}) {\figg{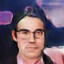}};
    \node[fig] at ({\figurewidth*4},{-\figureheight*5}) {\figg{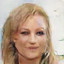}};
    \node[fig] at ({\figurewidth*5},{-\figureheight*5}) {\figg{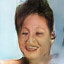}};
    \node[fig] at ({\figurewidth*6},{-\figureheight*5}) {\figg{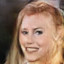}};
    \node[fig] at ({\figurewidth*7},{-\figureheight*5}) {\figg{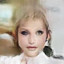}};
    \node[fig] at ({\figurewidth*8},{-\figureheight*5}) {\figg{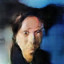}};
    \node[fig] at ({\figurewidth*9},{-\figureheight*5}) {\figg{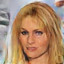}};
    
  \end{tikzpicture}
  \caption{AGE random samples (\CelebA) at $64{\times}64$ resolution.}
\end{figure}

\begin{figure}[!h]
  \centering\scriptsize
  \setlength{\figurewidth}{.09\textwidth}
  \setlength{\figureheight}{\figurewidth}
  \begin{tikzpicture}

    \tikzstyle{fig} = [draw=white,minimum size=\figurewidth,inner sep=0pt]

    \newcommand{\figg}[1]{\includegraphics[width=.97\figurewidth]{./supplement/ali_random_64x64/#1.jpg}};

    \node[fig] at ({\figurewidth*0},{-\figureheight*0}) {\figg{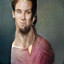}};
    \node[fig] at ({\figurewidth*1},{-\figureheight*0}) {\figg{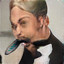}};
    \node[fig] at ({\figurewidth*2},{-\figureheight*0}) {\figg{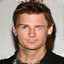}};
    \node[fig] at ({\figurewidth*3},{-\figureheight*0}) {\figg{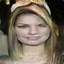}};
    \node[fig] at ({\figurewidth*4},{-\figureheight*0}) {\figg{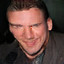}};
    \node[fig] at ({\figurewidth*5},{-\figureheight*0}) {\figg{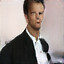}};
    \node[fig] at ({\figurewidth*6},{-\figureheight*0}) {\figg{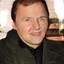}};
    \node[fig] at ({\figurewidth*7},{-\figureheight*0}) {\figg{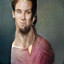}};
    \node[fig] at ({\figurewidth*8},{-\figureheight*0}) {\figg{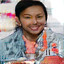}};
    \node[fig] at ({\figurewidth*9},{-\figureheight*0}) {\figg{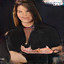}};
    \node[fig] at ({\figurewidth*0},{-\figureheight*1}) {\figg{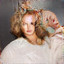}};
    \node[fig] at ({\figurewidth*1},{-\figureheight*1}) {\figg{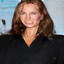}};
    \node[fig] at ({\figurewidth*2},{-\figureheight*1}) {\figg{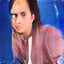}};
    \node[fig] at ({\figurewidth*3},{-\figureheight*1}) {\figg{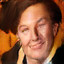}};
    \node[fig] at ({\figurewidth*4},{-\figureheight*1}) {\figg{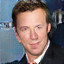}};
    \node[fig] at ({\figurewidth*5},{-\figureheight*1}) {\figg{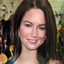}};
    \node[fig] at ({\figurewidth*6},{-\figureheight*1}) {\figg{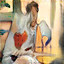}};
    \node[fig] at ({\figurewidth*7},{-\figureheight*1}) {\figg{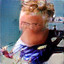}};
    \node[fig] at ({\figurewidth*8},{-\figureheight*1}) {\figg{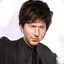}};
    \node[fig] at ({\figurewidth*9},{-\figureheight*1}) {\figg{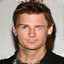}};
    \node[fig] at ({\figurewidth*0},{-\figureheight*2}) {\figg{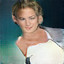}};
    \node[fig] at ({\figurewidth*1},{-\figureheight*2}) {\figg{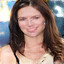}};
    \node[fig] at ({\figurewidth*2},{-\figureheight*2}) {\figg{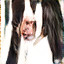}};
    \node[fig] at ({\figurewidth*3},{-\figureheight*2}) {\figg{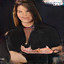}};
    \node[fig] at ({\figurewidth*4},{-\figureheight*2}) {\figg{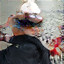}};
    \node[fig] at ({\figurewidth*5},{-\figureheight*2}) {\figg{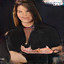}};
    \node[fig] at ({\figurewidth*6},{-\figureheight*2}) {\figg{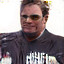}};
    \node[fig] at ({\figurewidth*7},{-\figureheight*2}) {\figg{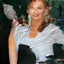}};
    \node[fig] at ({\figurewidth*8},{-\figureheight*2}) {\figg{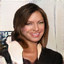}};
    \node[fig] at ({\figurewidth*9},{-\figureheight*2}) {\figg{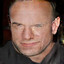}};
    \node[fig] at ({\figurewidth*0},{-\figureheight*3}) {\figg{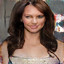}};
    \node[fig] at ({\figurewidth*1},{-\figureheight*3}) {\figg{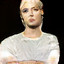}};
    \node[fig] at ({\figurewidth*2},{-\figureheight*3}) {\figg{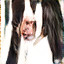}};
    \node[fig] at ({\figurewidth*3},{-\figureheight*3}) {\figg{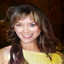}};
    \node[fig] at ({\figurewidth*4},{-\figureheight*3}) {\figg{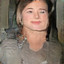}};
    \node[fig] at ({\figurewidth*5},{-\figureheight*3}) {\figg{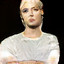}};
    \node[fig] at ({\figurewidth*6},{-\figureheight*3}) {\figg{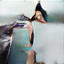}};
    \node[fig] at ({\figurewidth*7},{-\figureheight*3}) {\figg{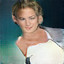}};
    \node[fig] at ({\figurewidth*8},{-\figureheight*3}) {\figg{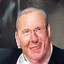}};
    \node[fig] at ({\figurewidth*9},{-\figureheight*3}) {\figg{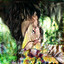}};
    \node[fig] at ({\figurewidth*0},{-\figureheight*4}) {\figg{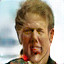}};
    \node[fig] at ({\figurewidth*1},{-\figureheight*4}) {\figg{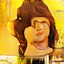}};
    \node[fig] at ({\figurewidth*2},{-\figureheight*4}) {\figg{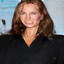}};
    \node[fig] at ({\figurewidth*3},{-\figureheight*4}) {\figg{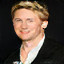}};
    \node[fig] at ({\figurewidth*4},{-\figureheight*4}) {\figg{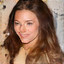}};
    \node[fig] at ({\figurewidth*5},{-\figureheight*4}) {\figg{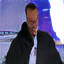}};
    \node[fig] at ({\figurewidth*6},{-\figureheight*4}) {\figg{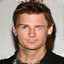}};
    \node[fig] at ({\figurewidth*7},{-\figureheight*4}) {\figg{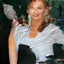}};
    \node[fig] at ({\figurewidth*8},{-\figureheight*4}) {\figg{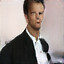}};
    \node[fig] at ({\figurewidth*9},{-\figureheight*4}) {\figg{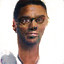}};
    \node[fig] at ({\figurewidth*0},{-\figureheight*5}) {\figg{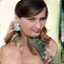}};
    \node[fig] at ({\figurewidth*1},{-\figureheight*5}) {\figg{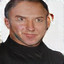}};
    \node[fig] at ({\figurewidth*2},{-\figureheight*5}) {\figg{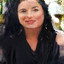}};
    \node[fig] at ({\figurewidth*3},{-\figureheight*5}) {\figg{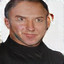}};
    \node[fig] at ({\figurewidth*4},{-\figureheight*5}) {\figg{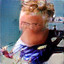}};
    \node[fig] at ({\figurewidth*5},{-\figureheight*5}) {\figg{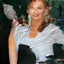}};
    \node[fig] at ({\figurewidth*6},{-\figureheight*5}) {\figg{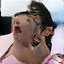}};
    \node[fig] at ({\figurewidth*7},{-\figureheight*5}) {\figg{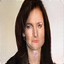}};
    \node[fig] at ({\figurewidth*8},{-\figureheight*5}) {\figg{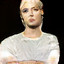}};
    \node[fig] at ({\figurewidth*9},{-\figureheight*5}) {\figg{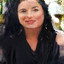}};

  \end{tikzpicture}
  \caption{ALI random samples (\CelebA) at $64{\times}64$ resolution.}
\end{figure}

\begin{figure}[!h]
  \centering\scriptsize
  \setlength{\figurewidth}{.09\textwidth}
  \setlength{\figureheight}{\figurewidth}
  \begin{tikzpicture}

    \tikzstyle{fig} = [draw=white,minimum size=\figurewidth,inner sep=0pt]

    \newcommand{\figg}[1]{\includegraphics[width=.97\figurewidth]{./supplement/pine_random_64x64/#1.jpg}};

    \node[fig] at ({\figurewidth*0},{-\figureheight*0}) {\figg{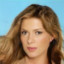}};
    \node[fig] at ({\figurewidth*1},{-\figureheight*0}) {\figg{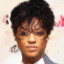}};
    \node[fig] at ({\figurewidth*2},{-\figureheight*0}) {\figg{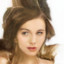}};
    \node[fig] at ({\figurewidth*3},{-\figureheight*0}) {\figg{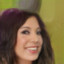}};
    \node[fig] at ({\figurewidth*4},{-\figureheight*0}) {\figg{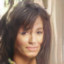}};
    \node[fig] at ({\figurewidth*5},{-\figureheight*0}) {\figg{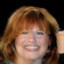}};
    \node[fig] at ({\figurewidth*6},{-\figureheight*0}) {\figg{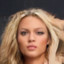}};
    \node[fig] at ({\figurewidth*7},{-\figureheight*0}) {\figg{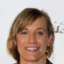}};
    \node[fig] at ({\figurewidth*8},{-\figureheight*0}) {\figg{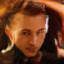}};
    \node[fig] at ({\figurewidth*9},{-\figureheight*0}) {\figg{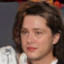}};
    \node[fig] at ({\figurewidth*0},{-\figureheight*1}) {\figg{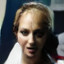}};
    \node[fig] at ({\figurewidth*1},{-\figureheight*1}) {\figg{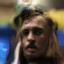}};
    \node[fig] at ({\figurewidth*2},{-\figureheight*1}) {\figg{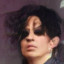}};
    \node[fig] at ({\figurewidth*3},{-\figureheight*1}) {\figg{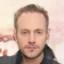}};
    \node[fig] at ({\figurewidth*4},{-\figureheight*1}) {\figg{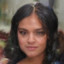}};
    \node[fig] at ({\figurewidth*5},{-\figureheight*1}) {\figg{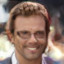}};
    \node[fig] at ({\figurewidth*6},{-\figureheight*1}) {\figg{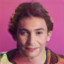}};
    \node[fig] at ({\figurewidth*7},{-\figureheight*1}) {\figg{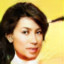}};
    \node[fig] at ({\figurewidth*8},{-\figureheight*1}) {\figg{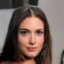}};
    \node[fig] at ({\figurewidth*9},{-\figureheight*1}) {\figg{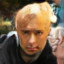}};
    \node[fig] at ({\figurewidth*0},{-\figureheight*2}) {\figg{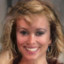}};
    \node[fig] at ({\figurewidth*1},{-\figureheight*2}) {\figg{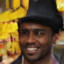}};
    \node[fig] at ({\figurewidth*2},{-\figureheight*2}) {\figg{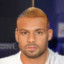}};
    \node[fig] at ({\figurewidth*3},{-\figureheight*2}) {\figg{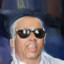}};
    \node[fig] at ({\figurewidth*4},{-\figureheight*2}) {\figg{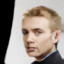}};
    \node[fig] at ({\figurewidth*5},{-\figureheight*2}) {\figg{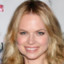}};
    \node[fig] at ({\figurewidth*6},{-\figureheight*2}) {\figg{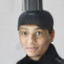}};
    \node[fig] at ({\figurewidth*7},{-\figureheight*2}) {\figg{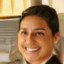}};
    \node[fig] at ({\figurewidth*8},{-\figureheight*2}) {\figg{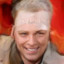}};
    \node[fig] at ({\figurewidth*9},{-\figureheight*2}) {\figg{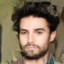}};
    \node[fig] at ({\figurewidth*0},{-\figureheight*3}) {\figg{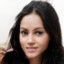}};
    \node[fig] at ({\figurewidth*1},{-\figureheight*3}) {\figg{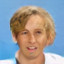}};
    \node[fig] at ({\figurewidth*2},{-\figureheight*3}) {\figg{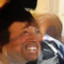}};
    \node[fig] at ({\figurewidth*3},{-\figureheight*3}) {\figg{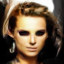}};
    \node[fig] at ({\figurewidth*4},{-\figureheight*3}) {\figg{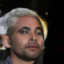}};
    \node[fig] at ({\figurewidth*5},{-\figureheight*3}) {\figg{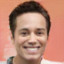}};
    \node[fig] at ({\figurewidth*6},{-\figureheight*3}) {\figg{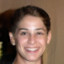}};
    \node[fig] at ({\figurewidth*7},{-\figureheight*3}) {\figg{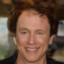}};
    \node[fig] at ({\figurewidth*8},{-\figureheight*3}) {\figg{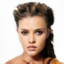}};
    \node[fig] at ({\figurewidth*9},{-\figureheight*3}) {\figg{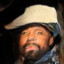}};
    \node[fig] at ({\figurewidth*0},{-\figureheight*4}) {\figg{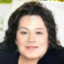}};
    \node[fig] at ({\figurewidth*1},{-\figureheight*4}) {\figg{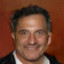}};
    \node[fig] at ({\figurewidth*2},{-\figureheight*4}) {\figg{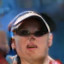}};
    \node[fig] at ({\figurewidth*3},{-\figureheight*4}) {\figg{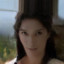}};
    \node[fig] at ({\figurewidth*4},{-\figureheight*4}) {\figg{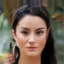}};
    \node[fig] at ({\figurewidth*5},{-\figureheight*4}) {\figg{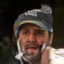}};
    \node[fig] at ({\figurewidth*6},{-\figureheight*4}) {\figg{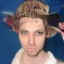}};
    \node[fig] at ({\figurewidth*7},{-\figureheight*4}) {\figg{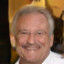}};
    \node[fig] at ({\figurewidth*8},{-\figureheight*4}) {\figg{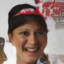}};
    \node[fig] at ({\figurewidth*9},{-\figureheight*4}) {\figg{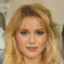}};
    \node[fig] at ({\figurewidth*0},{-\figureheight*5}) {\figg{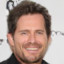}};
    \node[fig] at ({\figurewidth*1},{-\figureheight*5}) {\figg{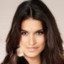}};
    \node[fig] at ({\figurewidth*2},{-\figureheight*5}) {\figg{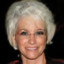}};
    \node[fig] at ({\figurewidth*3},{-\figureheight*5}) {\figg{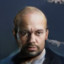}};
    \node[fig] at ({\figurewidth*4},{-\figureheight*5}) {\figg{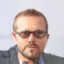}};
    \node[fig] at ({\figurewidth*5},{-\figureheight*5}) {\figg{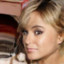}};
    \node[fig] at ({\figurewidth*6},{-\figureheight*5}) {\figg{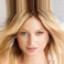}};
    \node[fig] at ({\figurewidth*7},{-\figureheight*5}) {\figg{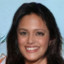}};
    \node[fig] at ({\figurewidth*8},{-\figureheight*5}) {\figg{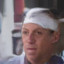}};
    \node[fig] at ({\figurewidth*9},{-\figureheight*5}) {\figg{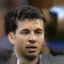}};

  \end{tikzpicture}
  \caption{\PIONEER random samples (\CelebA) at $64{\times}64$ resolution.}
\end{figure}

\end{document}